\documentclass{ecai}    % camerat-ready版本
\usepackage{latexsym}
\usepackage{amssymb}
\usepackage{amsmath}
\usepackage{amsthm}
\usepackage{booktabs}
\usepackage{enumitem}
\usepackage{graphicx}
\usepackage{color}

\usepackage{subfigure}
\usepackage{mathtools}
\usepackage{xcolor}
\usepackage{array}
\usepackage{bbm}
\usepackage{multirow}
\usepackage{makecell}
\usepackage{graphicx}
\graphicspath{{/}}   %添加Figure文件夹路径
\usepackage{algorithm}        %虽然能解决算法冲突问题,但是先不使用
\usepackage{algorithmic}  

\usepackage{diagbox}
\usepackage{ulem}
\newcolumntype{L}[1]{>{\raggedright\let\newline\\\arraybackslash\hspace{0pt}}m{#1}}
\newcolumntype{C}[1]{>{\centering\let\newline\\\arraybackslash\hspace{0pt}}m{#1}}
\newcolumntype{R}[1]{>{\raggedleft\let\newline\\\arraybackslash\hspace{0pt}}m{#1}}

\newcommand{\BibTeX}{B\kern-.05em{\sc i\kern-.025em b}\kern-.08em\TeX}

\begin{document}

%%%%%%%%%%%%%%%%%%%%%%%%%%%%%%%%%%%%%%%%%%%%%%%%%%%%%%%%%%%%%%%%%%%%%%%%

\begin{frontmatter}

%%% Use this command to specify your submission number.
%%% In doubleblind mode, it will be printed on the first page.

\paperid{1390} 

%%% Use this command to specify the title of your paper.

\title{Adversarial Attack for Explanation Robustness of Rationalization Models}
% 对Rationalization模型解释鲁棒性的对抗攻击：一个实证研究
% 对于Rationalization模型的对抗攻击：解释鲁棒性的实证研究

%%% Use this combinations of commands to specify all authors of your 
%%% paper. Use \fnms{} and \snm{} to indicate everyone's first names 
%%% and surname. This will help the publisher with indexing the 
%%% proceedings. Please use a reasonable approximation in case your 
%%% name does not neatly split into "first names" and "surname".
%%% Specifying your ORCID digital identifier is optional. 
%%% Use the \thanks{} command to indicate one or more corresponding 
%%% authors and their email address(es). If so desired, you can specify
%%% author contributions using the \footnote{} command.
% 使用这组命令指定论文的所有作者。使用 \fnms{} 和 \snm{} 来指明每个人的名和姓。这将有助于出版商为论文集编制索引。如果您的名字不能整齐地分为 "名 "和 "姓",请使用合理的近似值。指定您的 ORCID 数字标识符是可选项。使用 \thanks{} 命令指明一位或多位通讯作者及其电子邮件地址。如果需要,您还可以使用 \footnote{} 命令指定作者的贡献。

\author[A]{\fnms{Yuankai}~\snm{Zhang}\footnote{Equal contribution.}}
\author[A]{\fnms{Lingxiao}~\snm{Kong}\footnotemark[1]}
\author[A]{\fnms{Haozhao}~\snm{Wang}\thanks{Corresponding Author. Email: hz\_wang@hust.edu.cn}}
\author[A]{\fnms{Ruixuan}~\snm{Li}\thanks{Corresponding Author. Email: rxli@hust.edu.cn}}
\author[B]{\fnms{Jun}~\snm{Wang}}
\author[A]{\fnms{Yuhua}~\snm{Li}}
\author[A]{\fnms{Wei}~\snm{Liu}}

\address[A]{School of Computer Science and Technology, Huazhong University of Science and Technology, China}
\address[B]{iWudao Tech}

%%% Use this environment to include an abstract of your paper.

% 不超过200个字符。目前188字符
\begin{abstract}
% Extractive rationale models have emerged as a prominent research area in eXplainable Artificial Intelligence (XAI) due to their capacity to offer explanations to users. These models establish a trust chain, termed as ``model-prediction-explanation-user'', by selecting a subset of input as rationale. While previous studies have highlighted the vulnerability of rationale models in prediction, their robustness in explanation has not been thoroughly investigated, which serves as another crucial link in the trust chain.
% In this paper, we categorize the robustness of rationale models into prediction robustness and explanation robustness. Then, we introduce \emph{UAT2E}, a variant of Universal Adversarial Triggers designed to attack the explanation robustness of rationale models. By employing gradient-based search on triggers and inserting them into the original input, we conduct adversarial attacks that induce significant changes in the rationale while maintaining its prediction.
% Experimental results on five datasets reveal the vulnerability of rationale models in terms of explanation. Subsequent to the attacks, rationale models tend to select more meaningless tokens and triggers, resulting in a shift in the selection of rationale. Consequently, we make a series of recommendations for improving rationale models in terms of explanation.
%
Rationalization models, which select a subset of input text as rationale—crucial for humans to understand and trust predictions—have recently emerged as a prominent research area in eXplainable Artificial Intelligence (XAI). However, most of previous studies mainly focus on improving the quality of the rationale, ignoring its robustness to malicious attack. Specifically, \textit{whether the rationalization models can still generate high-quality rationale under the adversarial attack remains unknown.} 
To explore this, this paper proposes \emph{UAT2E}, which aims to undermine the explainability of rationalization models without altering their predictions, thereby eliciting distrust in these models from human users.
\emph{UAT2E} employs the gradient-based search on triggers and then inserts them into the original input to conduct both the non-target and target attack.
% , inducing significant changes in the rationale while maintaining the prediction.
%
Experimental results on five datasets reveal the vulnerability of rationalization models in terms of explanation, where they tend to select more meaningless tokens under attacks. Based on this, we make a series of recommendations for improving rationalization models in terms of explanation.
\end{abstract}

\end{frontmatter}

%%%%%%%%%%%%%%%%%%%%%%%%%%%%%%%%%%%%%%%%%%%%%%%%%%%%%%%%%%%%%%%%%%%%%%%%

\section{Introduction}
\begin{figure}[t]
% \centering
\begin{minipage}{0.45\columnwidth}
\subfigure[Clean Input]{
    \label{fig:CleanAI} %% label for first subfigure
    \includegraphics[width=\textwidth]{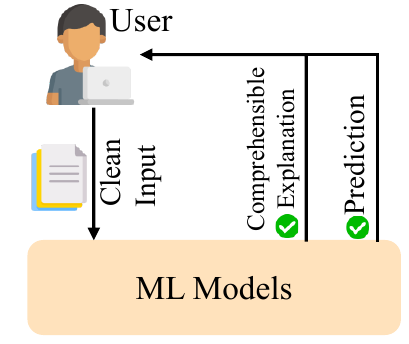}
}
\end{minipage}
\begin{minipage}{0.45\columnwidth}
\subfigure[Crafted Input]{
    \label{fig:AttackAI} %% label for first subfigure
    \includegraphics[width=\textwidth]{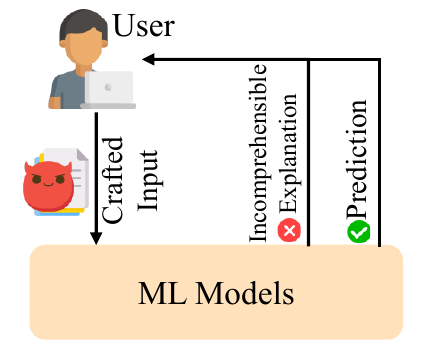}
}
\end{minipage}
% \vspace{-0.2cm}
\caption{Illustration of ML models with clean input and crafted input separately. (a) ML models not only returns correct prediction but also provides the comprehensible explanation to human user. (b) The explanation provided by ML models is incomprehensible for the crafted input.}
\label{fig:MainIdea} %% label for entire figure
\vspace{20pt}
\end{figure}

\begin{figure}[ht]
\centering
\includegraphics[width=0.99\columnwidth]{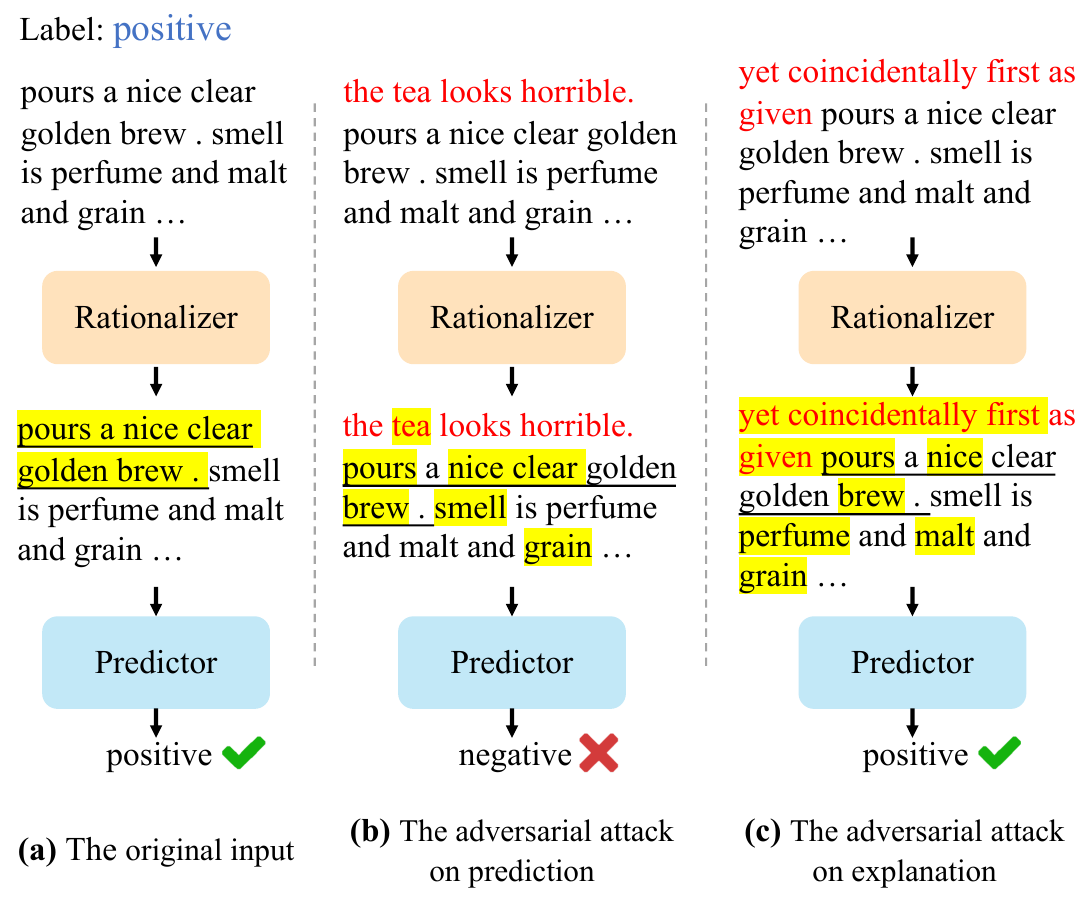}
% \vspace{-5pt}
\caption{(a) An example from a beer review sentiment classification dataset with correct prediction and rationale. (b) Inserting ``the tea looks horrible .'' causes the rationalizer to select ``tea'', ``smell'', and ``grain'', leading to an incorrect prediction. (c) Inserting ``yet coincidentally first as given'' results in maintaining a correct prediction but with an obviously incorrect rationale. The \uline{underline}, \textcolor{red}{red}, and \colorbox{yellow}{yellow} represent human-annotated rationale, triggers, and selected rationales, respectively.}
\label{fig_1: Prediction Robustness and Explanation Robustness}
\vspace{15pt}
\end{figure}

%whz
Explanation of deep learning models is the key to human \textit{comprehension} and \textit{trust} in their predictive outcomes by providing the corresponding explanations, as illustrated in Figure~\ref{fig:CleanAI}, which plays an important role in affecting whether these models can be applied to critical sectors such as finance and law.
Rationalization methods offer intrinsic justifications for model predictions by pinpointing salient evidence, emerging as a promising solution in the area of explainable artificial intelligence. As depicted in Figure~\ref{fig_1: Prediction Robustness and Explanation Robustness}, rationalization methods \citep{paranjape-etal-2020-information, 10.1145/3580305.3599299, lei-etal-2016-rationalizing} employ a rationalizer to extract a semantically coherent subset of the input text, known as the rationale. This rationale is intuitively recognized by humans as a decisive determinant of the subsequent predictor's output. \textit{By furnishing such interpretable rationales, rationalization methods significantly bolster human trust in the predictive outcomes.}

While existing studies have made great achievements in improving the explanation (i.e., rationale quality) of the rationalization models \citep{guerreiro-martins-2021-spectra, NEURIPS2022_2e0bd92a, NEURIPS2018_4c7a167b, liu-etal-2023-mgr, liu2023enhancing, wang2023pinto}, their explanation robustness to attack is rarely explored. Recently, \citet{chen-etal-2022-rationalization} exposed the prediction vulnerability of rationalization models by inserting crafted sentences into the original input text, leading to significant changes in predictions, as shown in Figure~\ref{fig_1: Prediction Robustness and Explanation Robustness}(b). Building upon this, some works \citep{Li_Hu_Chen_Xu_Tao_Zhang_2022, zhang-etal-2023-learning} employ adversarial training strategy to enhance the prediction robustness of these models, ensuring that the model predictions remain unchanged under adversarial attack. However, prior studies primarily focused on the robustness of rationalization methods concerning prediction while ignoring explanation. As illustrated in Figure~\ref{fig:AttackAI}, the models may provide incomprehensible explanation to human users when the input is crafted, leading to its reduced credibility. To this end, \textit{whether the explanation of rationalization models is robust to the adversarial attack remains mysterious.}

In this work, we investigate the explanation robustness of the rationalization models against adversarial attacks. More specifically, as illustrated in Figure \ref{fig_1: Prediction Robustness and Explanation Robustness}(c), we aim to craft and insert the attack trigger into the input text to noticeably change the rationale while keeping the prediction unchanged. In this way, the trust of human in the rationalization models can be significantly reduced.
We introduce UAT2E, a variant of Universal Adversarial Triggers \citep{wallace-etal-2019-universal}, which attacks explanations in non-target and target manner separately.\footnote{https://github.com/zhangyuankai2018/UAT2E} 
%%%%[多说一些实现的细节]
Specifically, UAT2E conducts the non-target attack by preventing the rationalizer from selecting the explainable tokens and conducts the target attack by limiting the rationalizer to only select the triggers.
To achieve this goal, we employs the mean squared error (MSE) loss to measure the difference in rationales and leverages the cross-entropy loss to calculate the difference in predictions. 
Then, according to the attack mode, UAT2E adaptively constructs label sequences to align the mismatched sequence lengths that result from inserting triggers. 
After that, UAT2E iteratively queries words from the vocabulary using a gradient-based approach and replaces tokens in the triggers to minimize the loss.

%%每个子图里加上(a)、(b)、(c)

%尽量不要加footnote
Based on the designed attack strategy, we investigate the explanation robustness of several typical and recent rationalization models on five public datasets, where the experimental results reveal several important findings (more findings can be found in Section~\ref{section: Main experiments}). \textit{First}, existing rationalization models are vulnerable to the attacks on explanation including both non-target and target attacks. \textit{Second}, the explanation vulnerablity of rationalization models arises from their inherent defects such as unmanageable sparsity, degeneration, and spurious correlation. \textit{Third}, using powerful encoders such as BERT and supervised training with human-annotated rationales in rationalization models does not guarantee the robustness of explanations; instead, it makes the explanation more susceptible to influence of attack. Based on these findings, we present a series of recommendations for improving explanation robustness of rationalization models. Although we mainly focus on the explanation robustness of rationalization models, we believe this work can provide a cautionary note regarding the robustness of all explainable machine learning systems. 

The contributions of this paper can be summarized as follows: 

\begin{itemize}
\item To the best of our knowledge, this work is the first to investigate the explanation robustness of the rationalization models.
\item We design UAT2E to conduct both non-target and target attacks over explanations of rationalization models. By employing gradient-based search to construct adversarial samples, UAT2E induces significant changes in rationale while maintaining prediction consistency.
\item We conduct extensive experiments on five public datasets, revealing the fragility of existing rationalization models in terms of explanation robustness and summarizing several reasons behind these phenomena. Additionally, we provide recommendations to enhance the explanation robustness of rationalization models.
\end{itemize}

%%%%%%%%%%%%%%%%%%%%%%%%%%%%%%%%%%%%%%%%%%%%%%%%%%%%%%%%%%%%%%%%%%%%%%%%

\section{Related Work} 
The rationalization model can be categorized into two types: extractive and generative. The extractive rationalization model \citep{guerreiro-martins-2021-spectra, NEURIPS2022_2e0bd92a, saha2023rationaleguided} involves selecting a subset from the original input to provide an explanation for the prediction. In contrast, the generative rationalization model \citep{NEURIPS2018_4c7a167b, wang2023pinto} uses text generation approaches to produce a piece of text explaining the prediction. While both approaches have their unique advantages, this paper focuses on related works on the robustness of the extractive rationalization model.

\noindent \textbf{The prediction robustness of rationalization model} Recent studies have focused on examining the prediction robustness of rationalization models. The prediction robustness refers to the model's ability to maintain its prediction unchanged when under attack.
\citet{chen-etal-2022-rationalization} explore the insertion of attack text into the original input by utilizing sentences from English Wikipedia or constructing them based on rules, inducing significant prediction flips. \citet{Li_Hu_Chen_Xu_Tao_Zhang_2022} employ TextAttack to modify specific words in the meaningless token regions of the original input, such as nouns, locations, numbers, and named entities, in order to generate adversarial samples. They also use adversarial training to enhance the prediction robustness of the rationalization model. \citet{zhang-etal-2023-learning} utilize the rationalization model as a defense strategy against adversarial attacks. They construct adversarial text using Glove and WordNet and employ adversarial training to ensure that the binary mask generated by the rationalizer effectively masks out the adversarial text, resulting in correct predictions. 
Unlike previous studies, we focus on the explanation robustness of rationalization models. The explanation robustness refers to the model's ability to maintain a consistent explanation when under attack. We focus on this by conducting attacks that induce significant changes in the explanation while maintaining the prediction.

\noindent \textbf{Degeneration and suprious correlations} Rationalization models face challenges, namely ``Degeneration'' \citep{yu-etal-2019-rethinking} and ``Spurious Correlation'' \citep{pmlr-v119-chang20c}. Degeneration occurs when the predictor overfits to noise generated by an undertrained generator, causing the generator to converge to a suboptimal model that selects meaningless tokens. Several approaches have been proposed to address the degeneration problem. \citet{yu-etal-2019-rethinking} introduced adversarial games and produces both positive and negative rationales. \citet{NEURIPS2022_2e0bd92a} employed a unified encoder between the generator and predictor. \citet{10.1145/3580305.3599299} assigned asymmetric learning rates to the two modules. The issue of spurious correlation arises because the maximum mutual information criterion can be influenced by false features associated with causal rationales or target labels, leading the generator to select content with false correlations. Existing works have attempted to address this problem from different perspectives, such as adopting environmental risk minimization \citep{pmlr-v119-chang20c} or the minimum conditional dependency criterion \citep{NEURIPS2023_87e82678}.

\noindent \textbf{Adversarial attacks in NLP} Adversarial attack research has played a critical role in uncovering vulnerabilities in interpretable NLP models \citep{chen2023darkexplanationspoisoningrecommender}. Adversarial attack methods can be categorized based on the input perturbations, including sentence-level \citep{jia-liang-2017-adversarial}, word-level \citep{wallace-etal-2019-universal, Herel_2023}, character-level \citep{ebrahimi-etal-2018-hotflip}, and embedding-level \citep{Li_Qiu_2021} attacks.
In this study, we use Universal Adversarial Triggers \citep{wallace-etal-2019-universal} to identify triggers in a white-box setting. By utilizing gradient-based search, we successfully identify the optimal combination of attack triggers.

\section{Problem Statement}
\label{section: Problem Statement}
\noindent \textbf{Notation} We denote the dataset by $\mathcal{D}=\{(x,y)\}$, where the input $x=x_1,x_2,...,x_T$ consists of $T$ sentences and each sentence $x_i=(x_{i,1},x_{i,2},...x_{i,n_i})$ contains $n_i$ tokens with $y$ referring to the sentence label. 

% The input $x$ is passed through the word embedding layer to obtain word vectors, denoted as $e = E(x)$. The weight of the embedding layer is represented by $\theta_{E} \in \mathbb{R}^{\mathcal{V} \times d}$, where $\mathcal{V}$ and $d$ denote the vocabulary size and word vector dimension, respectively.

\noindent \textbf{Extractive rationalization model} As shown in Figure~\ref{fig_1: Prediction Robustness and Explanation Robustness}(a), a typical extractive rationalization model comprises two components, i.e., a rationalizer and a predictor, where the rationalizer selects the rationale and the predictor makes the prediction. For each input $x$, the rationalizer first uses the word embedding layers to map it into vector $e=E(\theta_{E};x)$ where $\theta_{E} \in \mathbb{R}^{|\mathcal{V}| \times d}$ denotes its parameters. Then, the rationalizer adopts the Gumbel-Softmax reparameterization \citep{jang2017categorical} to sample and generate a discrete binary mask, $m=R(e)=(m_1,m_2,...,m_L)\in \{0,1\}^L$, from a Bernoulli distribution, where $L=\sum^{T}_{i=1}n_i$ for token-level rationale or $L=T$ for sentence-level rationale. In specific, the $i$-th element $m_i$ corresponds to sentence $x_i$, and thus $m_i$ will be extended according to the length $n_i$ for sentence-level. To this end, the rationale is calculated using $z=m \odot e$ equaling to a subset from the input. 

The predictor module $\hat{y}=C(z)$ makes a prediction $\hat{y}$ based on the rationale $z$. The overall prediction process can be defined as $M(x)=C(R(E(x))\odot E(x))$. Rationalization models are typically trained in an end-to-end fashion, where the cross-entropy loss between predictions and labels serves as the supervised signal. In this process, the rationale $z$ is generated unsupervised through the application of sparsity regularization.\footnote{Although coherent regularization is effective, we do not consider continuity constraints in order to compare each models as fairly as possible.} Appendix \ref{appendix: Model Detials} provides illustrations of the specific forms of sparsity regularization used in other models.

\noindent \textbf{Attack of rationalization model} We define triggers $a=(a_1,a_2,...,a_K)$ as input-agnostic sequences of tokens that, when inserted into any input from a dataset, cause significant changes in the rationale while maintaining the prediction. Typically, triggers consist of $K$ subsequences. For ease of implementation, each subsequence of triggers, $a_j=(a_{j,1},a_{j,2},...,a_{j,n_{a}})$, has the same length $n_{a}$. The attack $A(x,a,p)$ modifies the input $x$ by inserting triggers $a$ at specified positions $p=(p_1,p_2,...,p_K)$. The purpose is to ensure that the inserted triggers do not alter the semantics of individual sentence. The length of the adversarial sample $x_{adv}=A(x,a,p)$ is $L_{adv}=\sum^{T}_{i=1}n_i+K\times n_a$ for token-level rationale or $L_{adv}=T+K$ for sentence-level rationale.

\begin{table}[t]       
    \centering
    \caption{Notations of the raw input and the adversarial sample.}  
    \renewcommand{\arraystretch}{1.2}
    \label{Variables Definition}
    % \vspace{-15pt}
    \begin{center}
        \resizebox{0.99\columnwidth}{!}{
            \begin{tabular}{c|p{2.5cm}<{\centering}|p{3.3cm}<{\centering}}%p{2.5cm}<{\raggedright}}
                \hline
                 & $x$ & $x_{adv}=A(x,a,p)$ \\
                \hline
                Embedding & \makecell[c]{$e_{x}=E(x)$} & \makecell[c]{$e_{adv}=E(x_{adv})$}\\
                \hline
                Mask & \makecell[c]{$m_{x}=R(e_{x})$} & \makecell[c]{$m_{adv}=R(e_{adv})$}\\
                \hline
                Rationale & \makecell[c]{$z_{x}=m_{x}\odot e_{x}$} & \makecell[c]{$z_{adv}=m_{adv}\odot e_{adv}$} \\
                \hline
                Prediction & \makecell[c]{$\hat{y}_{x}=C(z_{x})$} & \makecell[c]{$\hat{y}_{adv}=C(z_{adv})$} \\
                \hline
            \end{tabular}
        }
    \end{center}
    \vspace{-10pt}
\end{table}

In order to establish a clear distinction between the original input $x$ and the adversarial sample $x_{adv}$, we provide definitions for word embedding, mask, rationale, and prediction in Table \ref{Variables Definition}. It should be noted that the embedding for triggers is denoted as $e_{a}=E(a)=(e_{a,1},e_{a,2},...,e_{a,K \times n_a})$.

\noindent \textbf{Assumptions of the attack} Reasonable assumptions play a crucial role in the effective evaluation of rationalization models. In the attack process, we make the assumption of having \emph{white-box} access to a well-trained rationalization model. This access enables us to obtain the target model's structure, gradient, word embedding weight, and sparsity level. The attacks are conducted exclusively during the \emph{inference} stage and not during the training stage.

\section{Universal Adversarial Triggers to attack the explanations (UAT2E)}
\subsection{Attack Objective}
\label{section: Attack Objective}
The objective of our method is to attack the explanation robustness, specifically by \emph{identifying the optimal trigger $a^*$ that maximizes the difference in rationale while maintaining the prediction}.
\begin{equation}
\begin{array}{c}
a^* = \arg \mathop{max}\limits_{a} \mathbb{E}_{x}[\mathbb{D}_{z}(z_{adv},z_x) - \beta \cdot \mathbb{D}_{y}(\hat{y}_{adv},\hat{y}_x)]
\label{Attack_Objective_la_equation}
\end{array}
\end{equation}
where $\mathbb{D}_{z}(\cdot,\cdot)$ measures the difference between the rationales $z_{adv}$ and $z_x$, $\mathbb{D}_{y}(\cdot,\cdot)$ measures the difference between the predictions $\hat{y}_{adv}$ and $\hat{y}_x$, and $\beta$ serves as the Lagrange multiplier. We employ the Mean Squared Error (MSE) loss to calculate the difference $\mathbb{D}_{z}(\cdot,\cdot)$ and cross-entropy loss to compute the difference $\mathbb{D}_{y}(\cdot,\cdot)$. $\beta$ is set to a value greater than $0$. 

\begin{figure}[t]
\centering
\includegraphics[width=0.99\columnwidth]{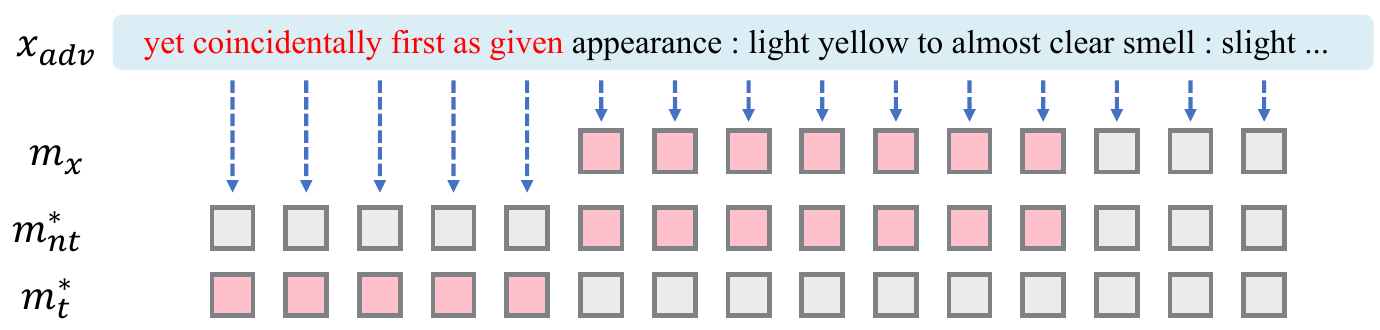}
% \vspace{-5pt}
\caption{Examples of label sequences under non-target and target attacks. Attack triggers are highlighted in \textcolor{red}{red}. Grey indicates 0, and pink indicates 1.}
\label{fig_2: non_target_and_target_sequence}
\vspace{15pt}
\end{figure}

\subsection{Non-target and Target Attack}
\label{section: Non-target and Target Attack}   

Equation (\ref{Attack_Objective_la_equation}) illustrates our intent, but inserting attack triggers leads to a mismatch in the lengths of $z_{adv}$ and $z_{x}$. To address this, we construct a label sequence $m^{*}$ from the discrete binary mask $m_{x}$ to replace $z_x$ and align the length of $m_{adv}$. We then calculate the difference between $m_{adv}$ and the label sequence $m^{*}$, as shown below:

\begin{equation}   
\label{Attack_Objective_la_equation_mask}      
\begin{array}{c}
\mathop{max}\limits_{a} \mathbb{E}_{x}[\mathbb{D}_{z}(m_{adv},m^{*})-\beta\cdot \mathbb{D}_{y}(\hat{y}_{adv},\hat{y}_x)]
\end{array}
\end{equation}
Similar to general adversarial attack methods \citep{7958570}, we consider the non-target attack $\mathcal{M}_{nt}$ and target attack $\mathcal{M}_{t}$.

\noindent \textbf{Non-target attack $\mathcal{M}_{nt}$} The goal of the non-target attack is to prevent the rationalizer module from selecting tokens previously chosen as rationale. This results in rationalization models selecting attack triggers or previously unselected tokens. To generate the label sequence for the non-target attack, denoted as $m^{*}_{nt}=\mathcal{M}_{nt}(m_{x})$, we insert $K$ $0$-sequences of length $n_{a}$ into the mask $m_{x}$ at specified positions $p$, as depicted by $m^{*}_{nt}$ in Figure \ref{fig_2: non_target_and_target_sequence}. Noting that in the non-target attack mode, the calculation of the difference is limited to the original input segment, totaling $L$ tokens. Furthermore, we adjust Equation (\ref{Attack_Objective_la_equation_mask}) by replacing ``maximize'' with ``minimize'' to align the optimization process with the gradient descent method.

\begin{equation}
\begin{array}{c}
\mathop{min}\limits_{a} \mathbb{E}_{x}[-\mathbb {D}_{z}(m_{adv},m^{*}_{nt})+\beta\cdot \mathbb{D}_{y}(\hat{y}_{adv},\hat{y}_x)]
\label{Non_target_Attack_Objective_la_equation}
\end{array}
\end{equation}

\noindent \textbf{Target attack $\mathcal{M}_{t}$} The goal of the target attack is to limit the rationalizer to selecting only attack triggers. The label sequence for the target attack is denoted as $m^{*}_{t}=\mathcal{M}_{t}(m_{x})$, where the elements corresponding to the triggers are assigned a value of 1, while other positions are set to 0, as depicted by $m^{*}_{t}$ in Figure \ref{fig_2: non_target_and_target_sequence}. Given the goal of the target attack, we need to minimize the difference between $m_{adv}$ and the label sequence $m^{*}_{t}$. The objective is shown as follows:

\begin{equation}
\begin{array}{c}
\mathop{min}\limits_{a} \mathbb{E}_{x}[\mathbb{D}_{z}(m_{adv},m^{*}_{t})+\beta\cdot \mathbb{D}_{y}(\hat{y}_{adv},\hat{y}_x)]
\label{Target_Attack_Objective_la_equation}
\end{array}
\end{equation}
By employing Equations (\ref{Non_target_Attack_Objective_la_equation}) and (\ref{Target_Attack_Objective_la_equation}), we can identify the optimal triggers $a^*$, through the standard gradient descent algorithm.

\subsection{Trigger Search Algorithm}
\label{section: Trigger Search Algorithm}
First, we initialize the attack triggers $a$ with the character at index 1 in the vocabulary $\mathcal{V}$. Next, we insert triggers into the original input and compute the loss based on Equation (\ref{Non_target_Attack_Objective_la_equation}) (or Equation (\ref{Target_Attack_Objective_la_equation})) depending on the attack mode. Then, we replace each token in the triggers with the one that minimizes the loss, using a greedy strategy. To determine the candidate token set, we use a KD\text{-}Tree to query the top-$k$ closest tokens by moving each token's embedding one step, sized $\eta$, in the gradient descent direction. We iteratively execute this process until we find the optimal trigger $a^*$ or reach the maximum number of search rounds $N$. More details of the attack process are in Appendix \ref{appendix: Attack Process}.

\section{Experiments}
We aim to explore the explanation robustness of existing models. Our experiments are conducted with five models, five datasets, two encoders, two training settings, and two attack modes, including a total of $200$ tests.

\subsection{Experimental setup}  
\noindent \textbf{Datasets} We consider five public datasets: Movie, FEVER, MultiRC from ERASER \citep{deyoung-etal-2020-eraser}, as well as Beer \citep{6413815} and Hotel \citep{10.1145/1835804.1835903}, two widely used datasets for rationalization. Such a setting encompasses both sentence-level and token-level rationalization tasks. More details about datasets are in Appendix \ref{appendix: Dataset Detials}.

\noindent \textbf{Models} We investigate five methods: RNP \citep{lei-etal-2016-rationalizing}, VIB \citep{paranjape-etal-2020-information}, SPECTRA \citep{guerreiro-martins-2021-spectra}, FR \citep{NEURIPS2022_2e0bd92a} and DR \citep{10.1145/3580305.3599299}. Details about these rationalization methods can be found in Appendix \ref{appendix: Model Detials}.

\noindent \textbf{Training details} All of the models are implemented using PyTorch and trained on a RTX3090 GPU. We adjust the sparsity parameter $\mathcal{S}$ based on the final sparsity level and select the model parameters based on the task performance achieved on the development dataset. Further training details can be found in Appendix \ref{appendix: More Training Details}.

\noindent \textbf{Attack details} We set the maximum length of adversarial samples $L_{adv}$ to 256 and the maximum number of search rounds $N$ to 100. We specified 5 insertion positions $p = (0, 2, 4, 6, -1)$, where ``$-1$'' represents the end position of the input. At each position, five tokens are inserted, denoted as $n_a=5$. The initial index of triggers is set to 1. For each trigger token, we query the 15 nearest candidate tokens using a KD-tree. The attack process employs an early stopping strategy: if the triggers no longer change after 10 epochs, the search is stopped. The step size $\eta$ is set to 1e4, and $\beta$ is set to 0.9. Experimental results are averaged across 5 random seeds.

\begin{figure*}[h]
    \centering
    \includegraphics[width=1.99\columnwidth]{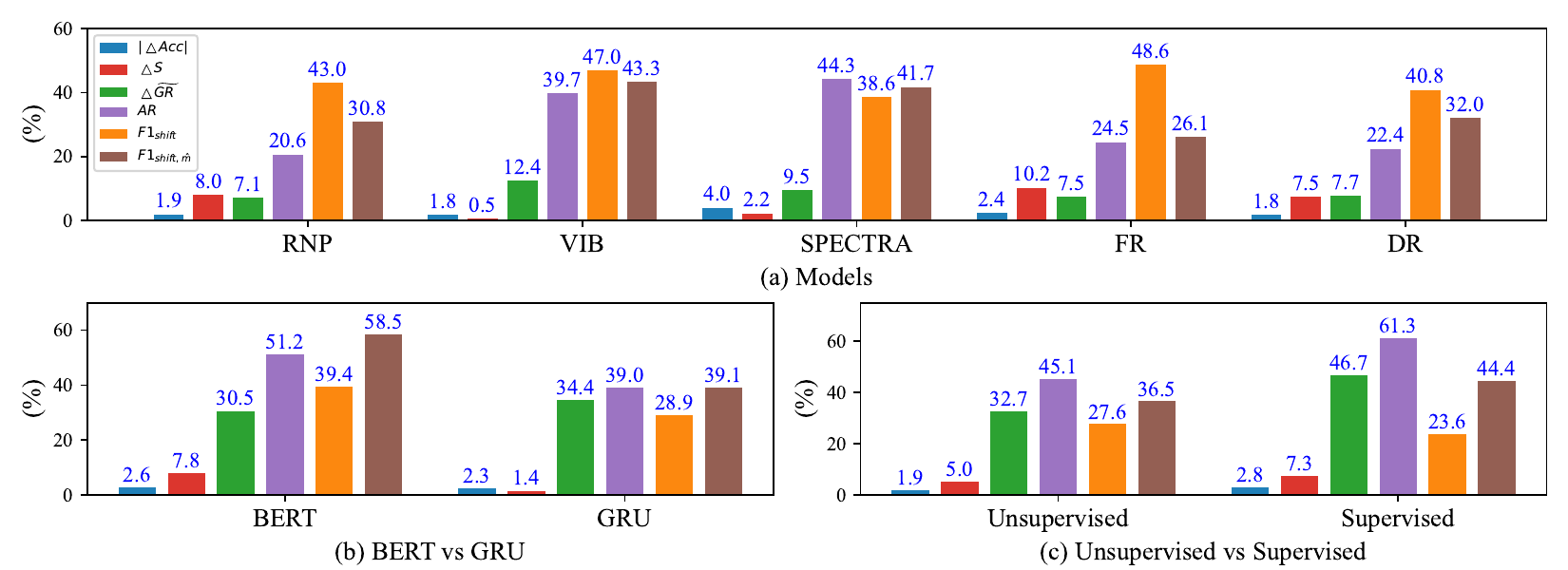}
    % \vspace{-10pt}
    \caption{Comparison across different settings. We compare three settings: (a) different models, (b) using BERT or GRU as an encoder, and (c) unsupervised training and supervised training with human-annotated rationales. The comparison is conducted using five models and five datasets.}
    \label{fig: Comparison across different settings}
    \vspace{10pt}
\end{figure*}

% 攻击之后产生degeneration和虚假关联
\begin{table*}[h]            %数据集为横轴,方法为纵轴     
    % \vspace{-0pt} 
    % \renewcommand{\arraystretch}{1.5}   %设置每一行内容上下方的空间
    \setlength{\tabcolsep}{3pt}    %设置列之间的距离
    \centering
    \caption{Samples with degeneration or spurious correlations after attacks. We present samples from different datasets to illustrate cases of model degeneration or spurious correlations resulting from attacks, along with statistical figures for these issues within the impacted samples. Human annotated rationales are \uline{underlined}. Rationales before attacks, triggers and rationales after attacks are highlighted in \textcolor{blue}{blue}, \textcolor{red}{red} and \colorbox{yellow}{yellow}, respectively.}
    \label{pick_degeneration_suprious_correlation}
    % \vspace{-15pt}
    \begin{center}
        \resizebox{0.99\linewidth}{!}{
            \begin{tabular}{m{1.5cm}<{\centering}|m{1cm}<{\centering}|L{15cm}|m{1.2cm}<{\centering}|m{1.5cm}<{\centering}|m{1cm}<{\centering}}
                \hline
                Model & Dataset & \multicolumn{1}{c|}{Context} & \makecell{Prediction\\(Label)} & Issue & \makecell{Percentage\\(\%)}\\
                \hline 
                RNP & Beer & \colorbox{yellow}{ \textcolor{red}{miniseries}} \textcolor{red}{ruins ich smashwords \text{!}} \colorbox{yellow}{\uline{\textcolor{blue}{inviting}}} \uline{\textcolor{blue}{copper color} with a \textcolor{blue}{sticky}, \textcolor{blue}{if thin} \colorbox{yellow}{\uline{,}} head \colorbox{yellow}{.}} caramel \colorbox{yellow}{and} apricots in the nose \colorbox{yellow}{.} \textcolor{red}{pondered friar vito assassinated hitler} wheat, dried \colorbox{yellow}{apricots}, caramel \colorbox{yellow}{,} and a dry finish \colorbox{yellow}{.} some lingering banana \colorbox{yellow}{,} too \colorbox{yellow}{.} \textcolor{red}{cliff} \colorbox{yellow}{\textcolor{red}{dismounted}} \textcolor{red}{smashwordsSell board} the hops \colorbox{yellow}{provide} a late \colorbox{yellow}{,} but pronounced character. pretty tasty \colorbox{yellow}{.} \textcolor{red}{\#\#" hack humour rumbled} \colorbox{yellow}{\textcolor{red}{\text{!}}} i \colorbox{yellow}{wouldn't} say it's as \textcolor{blue}{outstanding} as \colorbox{yellow}{others} from southern tier \colorbox{yellow}{like} pumking and \colorbox{yellow}{\textcolor{blue}{creme}} \textcolor{blue}{brule}e stout \colorbox{yellow}{, though} \textcolor{red}{valorbadgor concluded fatigue} & \makecell{Positive\\(Positive)} & Degeneration & 57.0 \\ 
                \hline
                DR & Beer & \textcolor{red}{same however time as only} poured from bomber \textcolor{blue}{into a pint glass} as part of a \colorbox{yellow}{side-by-side tasting} with duck-rabbit baltic porter. \uline{a : \textcolor{blue}{brown-black} with a \textcolor{blue}{thin head}.} \textcolor{red}{came. however time but} s : \colorbox{yellow}{not \textcolor{blue}{much} of a nose} from this bottle. \textcolor{blue}{coffee and caramel} or toffee. \textcolor{red}{time as although time same} t/mf : \colorbox{yellow}{great taste} : silky \colorbox{yellow}{smooth} in the \textcolor{blue}{mouth}. malty with a toffee finish. \textcolor{red}{came} \colorbox{yellow}{\textcolor{red}{as}} \textcolor{red}{same another however} \textcolor{blue}{crisper} finish than expected. d : very \colorbox{yellow}{\textcolor{blue}{drinkable}} example of the style, \textcolor{blue}{good} flavor combined with lighter mf makes for great experience. \textcolor{red}{time though time. however} & \makecell{Positive\\(Positive)} & Spurious Correlation & 48.1 \\  \hline
                RNP & Hotel &\textcolor{red}{however though neither fact first} recently stayed 6 days there. staff was very friendly and helpful. \textcolor{red}{. time this same the} \colorbox{yellow}{\textcolor{blue}{rooms} were clean} and quiet. i would rate the \colorbox{yellow}{\textcolor{blue}{beds}} as a 5 - not the worst i 've slept in but not the best \textcolor{blue}{.} \textcolor{red}{the the the} \colorbox{yellow}{\textcolor{red}{the all}} \textcolor{blue}{\uline{location is}} \colorbox{yellow}{\textcolor{blue}{\uline{great}}} \colorbox{yellow}{. \textcolor{blue}{pics}} of the \colorbox{yellow}{room} are a little exaggerated. \textcolor{red}{time this came this the} \colorbox{yellow}{bathroom is tiny}, small \colorbox{yellow}{\textcolor{blue}{closet} and no dressers} ( in the standard room ) and the \colorbox{yellow}{elevator} is a joke. we usually \textcolor{blue}{walked down} from \textcolor{blue}{the 6th \colorbox{yellow}{floor}} \colorbox{yellow}{rather than} wait. if you are only interested in a place to sleep and \textcolor{blue}{shower}, this works fine. \textcolor{red}{the the as the this} & \makecell{Positive\\(Positive)} & Spurious Correlation & 68.4 \\  \hline
                FR & Hotel & \textcolor{red}{deployed stockholm !logic !} time and time again as i stand jaded by brutal service and hotel properties \colorbox{yellow}{,} a westin hotel restores my faith \colorbox{yellow}{\textcolor{blue}{.}} \uline{the \textcolor{blue}{downtown location is great}}, staff was \textcolor{blue}{excellent} and you can never \colorbox{yellow}{go} wrong with a heavenly bed - - the first \colorbox{yellow}{to} market and still best by my standards \textcolor{blue}{.} \textcolor{red}{mentIonal \colorbox{yellow}{!} able u edUisLieUsly \colorbox{yellow}{!} locals \colorbox{yellow}{!} situation viz ! ation can normalAble !} & \makecell{Positive\\(Positive)} & Degeneration & 66.6 \\ \hline
                VIB & MultiRC & Where did Mr. Steadman go to get the paper ? || The grocery store \textcolor{red}{only neither yet quickly putting} He read the telegram again. In desperation he went back to the long distance booth, but found the line still out of order, and a wire had come giving the details of the damage done by the storm. \textcolor{red}{so kept next given just} It would be several days before communication could be established. There was no help coming from headquarters, and from the wording of the telegram there seemed to be a reason for their not giving clear details. \textcolor{red}{addition another the the another} He must get a copy of the paper. \uline{\textcolor{blue}{Reluctantly he went to the printing office and made known his}} \uline{\textcolor{blue}{ errand.}} \textcolor{red}{it several especially yet almost} \uline{\textcolor{blue}{Mr. Driggs was delighted to give him the paper -- he had it some place, though he very}} \uline{\textcolor{blue}{ seldom opened any of his exchanges. He}} \uline{\textcolor{blue}{evidently bore Mr. Steadman no ill - will for his plain talk two weeks ago.}} ...... \colorbox{yellow}{Mr.Steadman remarked carelessly that there was an editorial in it to which his attention had been drawn, on hearing which} \colorbox{yellow}{ Mr. Driggs turned his head and winked at an imaginary} \colorbox{yellow}{accomplice. \textcolor{red}{another first first part another}} & \makecell{Negative\\(Negative)} & Spurious Correlation & 42.6 \\ \hline
                SPECTRA & FEVER & the first inauguration of bill clinton was in the united states. \colorbox{yellow}{\textcolor{red}{ran ! ored ‡ !}} \uline{\textcolor{blue}{the first inauguration of bill clinton}} \uline{\textcolor{blue}{as the 42nd president of the united states was held on january 20, 1993 on the west front of the united states capitol}} \uline{\textcolor{blue}{building in washington, d. c..}} the inauguration marked the commencement of the first four - year term of bill clinton as president and al gore as vice president. \textcolor{red}{boro navarrekia ! . wal} at of age time of his first inauguration, clinton was the third - youngest person to become president, and the first from the baby boomer generation. \colorbox{yellow}{\textcolor{red}{! ! ! ie !}} \textcolor{red}{) 2war lu ! sselUresAUtnantTern} & \makecell{Positive\\(Positive)} & Degeneration & 60.7\\ \hline
                \hline
            \end{tabular}	
        }
    \end{center}
\end{table*}

\begin{figure*}[h]
    \centering
    \includegraphics[width=1.99\columnwidth]{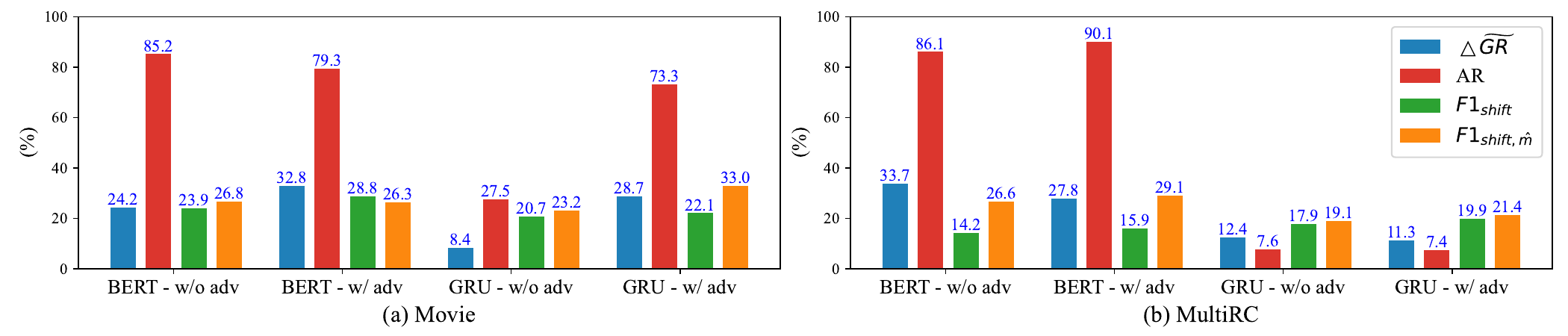}
    % \vspace{-10pt}
    \caption{Evaluating the impact of improving prediction robustness on explanation robustness. We train RNP on the Movie (a) and MultiRC datasets (b). ``w/o adv'' and ``w/ adv'' represent the cases without and with adversarial training, respectively.}
    \label{Figure: Enhance Prediction Robustness}
    \vspace{10pt}
\end{figure*}

\subsection{Evaluation Metric}
We evaluate the robustness of models in terms of \textit{task performance} and \textit{rationale quality}.

\noindent \textbf{Task performance} We compare the accuracy between the original and adversarial test sets. The absolute differences in accuracy are shown below:
\begin{equation*}   
\setlength{\abovedisplayskip}{3pt plus 2pt minus 8pt}
\setlength{\belowdisplayskip}{3pt plus 2pt minus 8pt}
|\triangle Acc| = \frac{1}{|\mathcal{D}|}\sum_{(x,y)\in \mathcal{D}}{|\mathbbm{1}_{[M(x)=y]} - \mathbbm{1}_{[M(A(x,a,p))=y]}|}
\end{equation*} 
Here, $|\mathcal{D}|$ represents the total number of samples, $| \cdot |$ denotes the absolute value, and $\mathbbm{1}$ is the indicator function. In our experiments, we considered a difference in accuracy within $5\%$ as the threshold, indicating that the predictions are generally consistent.

\noindent \textbf{Rationale quality} To evaluate the impact of the attack on rationales, we employ the following six metrics. Note that all metrics are averaged on the dataset consisting of non-flipped samples, which are defined as $\mathcal{D}_{nf}=\{x|M(x)=M(A(x,a,p))\}$. This is because a prediction flip will result in a more substantial change in rationale, leading to higher metric values. 
%The detailed description of the metrics mentioned in previous works is provided in Appendix \ref{appendix: Supplementary Description of Metrics}.

\begin{itemize}
    \item Sparsity ($\triangle S$): This metric calculates the ratio of selected tokens to all input tokens. The sparsity level varies before and after the attack due to the different degrees of sparsity in the label sequence under non-target and target attacks. Thus, we calculate sparsity's difference before and after the attack.
    \begin{equation*}
        \triangle S =  \frac{||m_{x}||}{L} - \frac{||m_{adv}||}{L_{adv}}
    \end{equation*}
    where $||\cdot||$ represents the $l_{1}$ norm.
    \item Gold Rationale F1 ($GR$): This metric assesses the F1 score between the rationale produced by the model and the human-annotated rationale. A decrease in GR typically indicates fragility in explanation robustness. 
    \begin{equation*}
        GR = \frac{||m \cap \hat{m}||}{||m \cup \hat{m}||}
    \end{equation*}
    Here, $\hat{m}$ represents the human-annotated mask, $m$ represents the mask generated by either the original input or the adversarial example. $\triangle GR = GR_{x} - GR_{adv}$, where $GR_{x}$ and $GR_{adv}$ denote the GR of the original input and the GR of the adversarial sample, respectively.
    \item $\triangle \widetilde{GR}$: This metric represents the relative difference of GR and is designed to facilitate the comparison of the impact on different models. 
    \begin{equation*}
        \triangle \widetilde{GR} = \frac{GR_{x}-GR_{adv}}{GR_{x}}
    \end{equation*}
    \item Attack Capture Rate (AR): AR represents the recall of attack triggers in the rationale generated by the adversarial sample. Models exhibiting strong explanation robustness should exclude attack triggers in their selections.
    \begin{equation*}
        AR = \frac{||m_{adv} \cap m^{*}_{t}||}{||m^{*}_{t}||}
        \end{equation*}
    \item Rationale Shift F1 ($F1_{shift}$): This metric assesses the F1 score between rationales before and after the attack, indicating the degree of token shifting.
    \begin{equation*}   
        F1_{shift} = \frac{||m_{adv} \cap m^{*}_{nt}||}{||m^{*}_{nt}||}
    \end{equation*}
    This metric assesses the consistency of the rationalizer's selection when the GR remains constant. If the rationalizer's selection shifts from one set of tokens to another, $F1_{shift}$ will decrease.
    \item Rationale Shift F1 on Annotation ($F1_{shift,\hat{m}}$): This metric evaluates the F1 score of the tokens selected by the model before and after the attack within the region of the human-annotated rationale.
    \begin{equation*}   
        F1_{shift,\hat{m}} = \frac{||m_{adv} \cap m^{*}_{nt} \cap \hat{m}||}{||m^{*}_{nt} \cap \hat{m}||}
    \end{equation*} 
    By comparing $F1_{shift}$ and $F1_{shift,\hat{m}}$, we can analyze the model's ability to recognize and retain tokens from the human-annotated rationale.
\end{itemize}

\subsection{Main experiments}
\label{section: Main experiments}
\noindent \textbf{Finding 1: Existing rationalization models exhibit significant fragility in explanation robustness} Figure \ref{fig: Comparison across different settings} (a) illustrates the vulnerability of rationalization models in terms of explanation robustness, even when predictions remain unchanged. It is worth noting that directly comparing different methods is meaningless due to various factors. For instance, VIB shows much smaller $\triangle S$ than other methods. But that does not mean it's more robust than other methods. The main reason is that it samples top-$k$ tokens while other methods sample rationales with Gumbel-Softmax. However, when sampling from the Gumbel-Softmax distribution, VIB's $\triangle S$ increases. Therefore, instead of horizontally comparing different methods, we focus on the individual performance of each method to highlight the widespread fragility in terms of explanation. Following UAT2E attacks, these models tend to shift the selection of rationales from tokens prior to the attack to meaningless tokens or triggers, resulting in a decrease in $\triangle \widetilde{GR}$, $F1_{shift}$, and $F1_{shift,\hat{m}}$, while increasing AR. 

\noindent \textbf{Finding 2: Rationalization models tend to exhibit degeneration or select spurious correlations when subjected to attacks}
Analysis of sample cases indicates that these shifts occur because UAT2E identifies trigger combinations leading the model to experience degeneration or select spurious correlations, as presented in Table \ref{pick_degeneration_suprious_correlation}. In the beer dataset, 57.0\% of the samples exhibit a higher tendency to select ``,'', ``.'', or other meaningless tokens after being attacked, while 48.1\% of the samples choose non-appearance-related content.

\noindent \textbf{Finding 3: Using a powerful encoder or supervised training with human-annotated rationales fails to mitigate degeneration and spurious correlations resulting from attacks} Compared to GRU, the model with BERT demonstrates higher $GR$ on sentence-level datasets and experiences a greater $GR$ boost through supervised training before attacks. When considering the rationale quality before and after attacks, BERT-based models generally experience greater impact, resulting in larger $\triangle GR$ values. However, the $\triangle \widetilde{GR}$ values are smaller due to the higher pre-attack $GR$. The disparity in sparsity indicates that BERT is more vulnerable to attacks, resulting in a more significant increase in sparsity compared to GRU. Consequently, precision and $GR$ experience a decrease, as illustrated in Figure \ref{fig: Comparison across different settings} (b). Models trained under supervision with human-annotated rationales fail to prevent the impact of attacks. Specifically, they tend to select more meaningless tokens and triggers while discarding the originally selected human-annotated rationale tokens. This also leads to increased sparsity and decreases in precision, recall, and the $GR$ score, as shown in Figure \ref{fig: Comparison across different settings} (c).

These two approaches help the model recognize rationales and provide more accurate gradient information, which assists UAT2E in selecting tokens that can induce model degradation or spurious correlations. Notably, the discrepancy between $F1_{shift}$ and $F1_{shift,\hat{m}}$ is more pronounced in both situations. This indicates that using the two approaches is better at identifying and preserving tokens annotated as rationales by humans, resulting in more shifts occurring on meaningless tokens.

\noindent \textbf{Finding 4: Enhancing prediction robustness does not effectively improve explanation robustness} We conduct experiments on the Movie and MultiRC datasets to investigate whether improving prediction robustness can enhance explanation robustness. Specifically, we train RNP on the Movie, MultiRC, Movie\_ADV, and MultiRC\_ADV datasets. The Movie\_ADV and MultiRC\_ADV datasets, mentioned in \citep{Li_Hu_Chen_Xu_Tao_Zhang_2022}, are used for adversarial training to enhance prediction robustness. Following the model training, we perform non-target attacks, and the experimental results are depicted in Figure \ref{Figure: Enhance Prediction Robustness}. It is noteworthy that we employ $\triangle \widetilde{GR}$ to intuitively compare the impact on models with different encoders. The results indicate that enhancing prediction robustness through adversarial training does not significantly improve explanation robustness. Particularly, RNP with GRU on the Movie dataset experiences a more pronounced impact after adversarial training.

\begin{figure}[h]
    \centering
    \includegraphics[width=0.99\columnwidth]{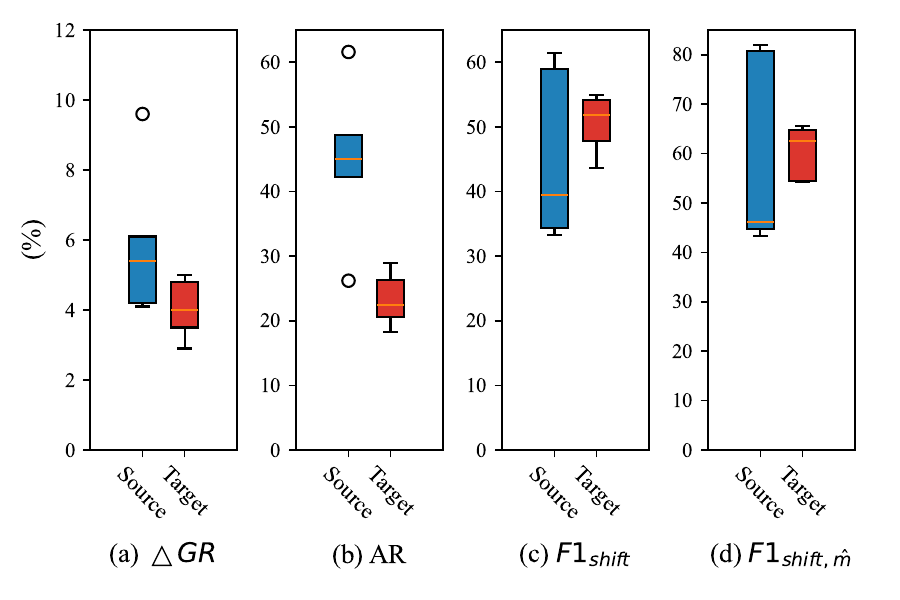}
    % \vspace{-5pt}
    \caption{Evaluating the transferability of identified triggers. We conduct tests on five models using BERT as the encoder, across three datasets: Beer, Movie, and FEVER. Triggers identified on one model (source model) are then transferred to other models (target models). \textit{Blue}: Triggers apply to the source model, \textit{Red}: Triggers are transferred to target models, and the experimental results are averaged.}
    \label{fig_3: transferability}
\end{figure}

\noindent \textbf{Finding 5: Utilizing gradient-based search in attacks to facilitate trigger selection} We conduct transferability tests using the identified triggers. Specifically, we transfer the triggers from a source model to a target model and assess the attack effects, as depicted in Figure \ref{fig_3: transferability}. It is worth noting that we do not have access to any information about the target model in order to demonstrate the effectiveness of trigger transfer in a black-box setting. Despite a slight reduction in effectiveness, the triggers are still capable of influencing the target model. However, the AR value is lower, which could potentially be attributed to the absence of a gradient-based search, making it challenging for the model to select triggers.

\subsection{Ablation Study} 
\begin{figure}[ht]
\centering
\includegraphics[width=0.99\columnwidth]{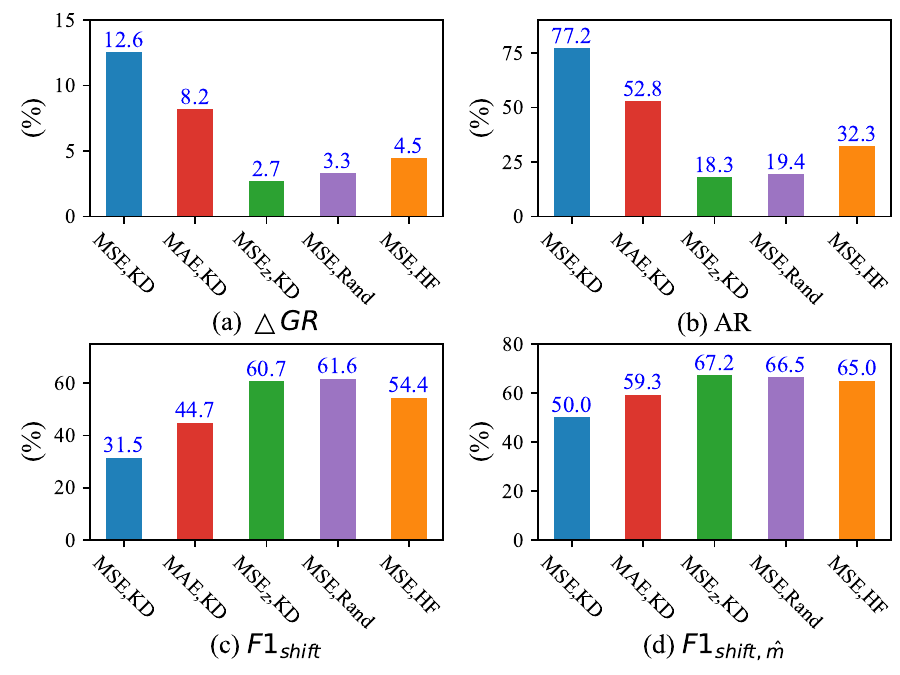}
% \vspace{-10pt}
\caption{The impact of measurement functions and query methods. We employed MSE and MAE as measurement functions. ``$MSE_{z}$'' was used to calculate the differences in rationale embeddings, specifically $MSE_{z}(m \odot e_{x}, m^{*} \odot e_{x})$. For querying candidate tokens, we uses the KD-Tree, random selection in each round, or the HotFlip \citep{ebrahimi-etal-2018-hotflip} method.}
\label{fig_2: ablation experiment}
\vspace{10pt}
\end{figure}

We conducted non-target attacks on VIB, utilizing GRU as the encoder, on three datasets: Hotel, Movie, and MultiRC, and the results are shown in Figure \ref{fig_2: ablation experiment}. 
When comparing Mean Absolute Error (MAE) with Mean Squared Error (MSE) as a rationale measurement function, the use of MSE yields a more significant attack effect. This discrepancy arises from MSE more prominently capturing the differences between $m^{adv}$ and the label sequence $m^*$. The attack effects of calculating MSE in the embedding space and executing attacks by randomly selecting words in each round are similar. This is because the former emphasizes distinctions between word vectors rather than masks, resulting in gradient fluctuations and a reduction in the attack effect. Additionally, when employing the HotFlip \citep{ebrahimi-etal-2018-hotflip} query method, the attack effect closely resembles that of randomly selecting words in each round.

\begin{figure}[ht]
    \centering
    \includegraphics[width=0.99\columnwidth]{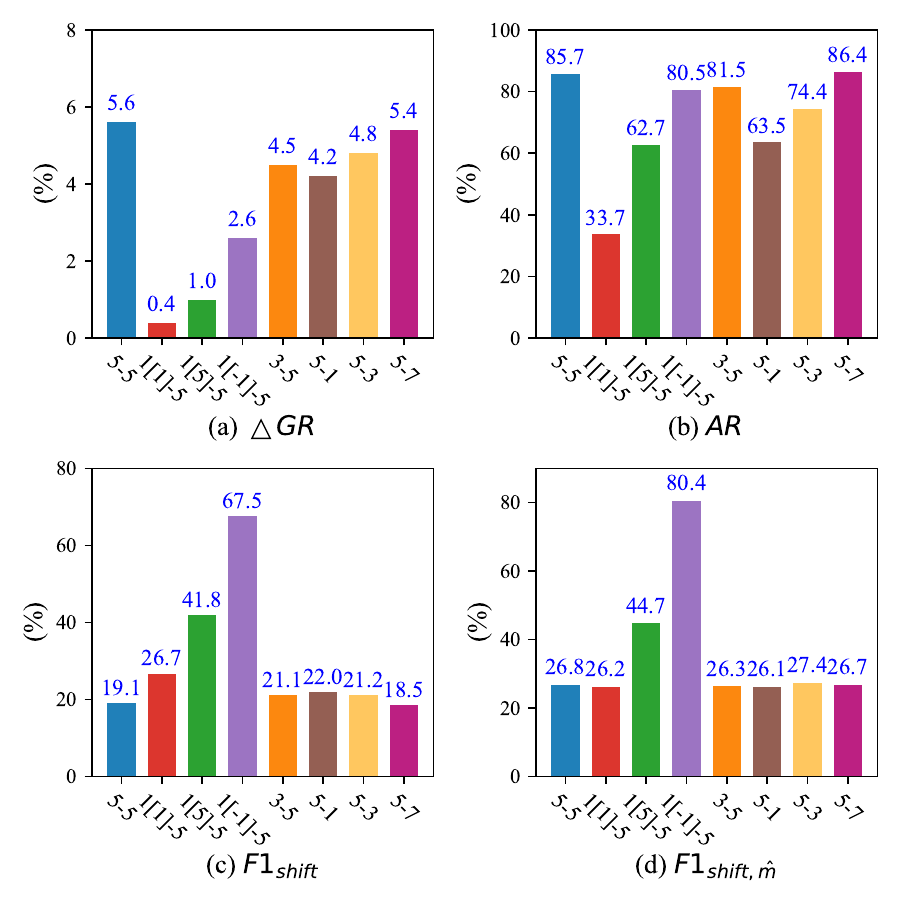}
    % \vspace{-5pt}
    \caption{Comparison various insertion positions and number of triggers. We conducted non-target attacks on RNP using the Movie and MultiRC datasets. ``\textit{5-5}'': insert 5 groups of triggers, each with 5 tokens. ``\textit{1[p]-5}'': insert 1 group of triggers before the $p$-th sentence. The ``-1'' position means insert 1 group of triggers at the end of the document. For the 5-group strategy, we specified positions as p=(0,2,4,6,-1), and for the 3-group strategy, p=(0,4,-1).}
    \label{Figure: Triggers Position And Number}
\vspace{10pt}
\end{figure}

\subsection{Triggers Position And Number} 
We examine the impact of different insertion positions and numbers of triggers on model performance, as depicted in Figure \ref{Figure: Triggers Position And Number}. Gradually increasing the number of trigger groups and tokens per group intensifies their influence on the model. Notably, the variation in the number of trigger sets exhibits a more significant effect on model performance. Additionally, triggers inserted near the end of the document have a greater impact. This observation aligns with previous findings by \citet{chen-etal-2022-rationalization} and can be attributed to the fact that rationale positions in Movie and MultiRC datasets are typically located close to the end of the document.

\section{Recommendations}
Based on our experimental results and analysis, we offer several recommendations to researchers and practitioners:

\noindent \textbf{Conducting rigorous evaluations of rationalization models} Researchers should assess both task performance and rationale quality by implementing various types of attacks on explanations and predictions. This examination helps determine whether rationalization models exhibit both high prediction and explanation robustness.

\noindent \textbf{Exploring defense mechanisms to enhance explanation robustness} Our experimental findings reveal rationalization models suffer from issues such as degeneration and spurious correlation after being attacked. Therefore, researchers should explore the development of defense mechanisms to protect rationalization models from attacks and reduce the occurrence of degeneration and spurious correlation.

\noindent \textbf{Establishing robustness evaluation benchmarks and metrics} It is imperative for researchers to construct benchmarks that facilitate standardized and rigorous evaluation of model robustness. Such benchmarks enable the identification of strengths and weaknesses across various models using unified criteria. Additionally, it is important to note that Gold Rationale F1 (GR) may not accurately reflect rationale shifts. Therefore, developing more effective evaluation metrics is essential for measuring these shifts accurately.

\section{Conclusion}        
In this study, we investigate the robustness of rationalization models in terms of explanation. To explore this, we propose UAT2E, a variant of Universal Adversarial Triggers. UAT2E attacks explanations in non-target and target manner separately, resulting in significant shifts in rationales while maintaining predictions.

Based on the experimental findings, it is evident that existing rationalization models generally exhibit vulnerabilities in explanation, making them susceptible to attacks that result in significant shifts in rationales. These vulnerabilities can be attributed to degeneration or spurious correlations after being attacked. Furthermore, despite employing techniques to improve rationale quality, such as using more powerful encoders or utilizing supervised training with human-annotated rationales, the explanation robustness of rationalization models does not significantly improve.

Based on our findings, we supplement our findings with a series of recommendations for enhancing the explanation robustness of rationalization models.

\section{Ethics Statement} 
The data and resources used in this study are publicly available and have been widely used in previous research. It is important to note that in our experiments, some triggers consist of email addresses, which are sourced from a vocabulary corpus dataset. This has the potential to result in the disclosure of personal privacy.

%%%%%%%%%%%%%%%%%%%%%%%%%%%%%%%%%%%%%%%%%%%%%%%%%%%%%%%%%%%%%%%%%%%%%%%%

%%% Use this environment to include acknowledgements (optional).
%%% This will be omitted in doubleblind mode.
% 致谢部分
\begin{ack}
We thank the anonymous reviewers for their constructive comments.
This work is supported by National Natural Science Foundation of China under grants 62376103, 62302184, 62206102, Science and Technology Support Program of Hubei Province under grant 2022BAA046, and CCF-AFSG Research Fund.
\end{ack}

%%%%%%%%%%%%%%%%%%%%%%%%%%%%%%%%%%%%%%%%%%%%%%%%%%%%%%%%%%%%%%%%%%%%%%%%

%%% Use this command to include your bibliography file.
% 包含引用
\bibliography{custom}

\clearpage

% 【appendix先整体编译,然后拆分出来放到支撑材料。】
\appendix
\onecolumn
\section{Details of Attack Process}
\label{appendix: Attack Process}
In this section, we introduce the details of the attack process with Algorithm \ref{algorithm: Attack Process}.

\noindent \textbf{Initializing attack triggers} We initialize each token in the attack triggers $a$ with the character at index 1 in the vocabulary $\mathcal{V}$.

\noindent \textbf{Loss calculation for adversarial dataset} For every input $x$ in the original dataset $\mathcal{D}$, we insert the corresponding triggers $a_i$ at each specified position $p_i$ to generate the adversarial dataset $\mathcal{D}_{a}$. The formula is as follows:
\begin{equation}
\begin{split}
\mathcal{D}_{a}&=\{(m_{x}, m_{adv}, \hat{y}_{x}, \hat{y}_{adv})|x\in \mathcal{D},x_{adv}=A(x,a,p)\}
\label{construct_D'}
\end{split}
\end{equation}

We construct the corresponding label sequence $m^*$ based on the attack mode $\mathcal{M}$ and compute the loss $l$.
\begin{equation}
\begin{array}{c}
l=
\begin{cases}
$Equation (\ref{Non_target_Attack_Objective_la_equation}) \text{,}$ \quad\ \mathcal{M}=\mathcal{M}_{nt} \\
$Equation (\ref{Target_Attack_Objective_la_equation}) \text{,}$ \quad\ \mathcal{M}=\mathcal{M}_{t} \\
\end{cases}
\label{loss_l}
\end{array}
\end{equation}

\noindent \textbf{Querying candidate tokens} We calculate the gradients $g$ for the triggers $a$. 
\begin{equation}
\begin{array}{c}
g = \frac{\partial{l}}{\partial{e_{a}}}=(g_{1},g_{2},...,g_{K \times n_a})\in \mathbb{R}^{(K \times n_{a}) \times d}
\label{calculate_gradient}
\end{array}
\end{equation}

Subsequently, utilizing the standard gradient descent algorithm with a step size of $\eta$, we generate a corresponding new vector $e_{a,i}'$ for each token $e_{a,i}$ in triggers $a$.
\begin{equation}
\begin{array}{c}
e_{a,i}'=e_{a,i}-\eta \cdot g_{i}, i=1, 2,..., ( K \times n_a )
\label{GD}
\end{array}
\end{equation}

To efficiently query the nearest $k$ candidate tokens, we construct a KD-tree, denoted as $\mathcal{T}$, using $\theta_{E}$. Here, $\theta_{E}$ is obtained from static word vectors, such as GloVe, or dynamic word vectors, such as BERT. The construction of KD-tree $\mathcal{T}$ is represented as follows:
\begin{equation}
\begin{array}{c}
\mathcal{T}=\text{KD-TREE}(\theta_{E})
\label{create_KDtree}
\end{array}
\end{equation}

By utilizing the KD-tree $\mathcal{T}$, we query the new vector $e_{a,i}'$ and obtain the indexes of the $k$ nearest neighbors to form a candidate set $\mathcal{I}_{e_{a,i}}$ for the trigger token $e_{a,i}$.
\begin{equation}
\setlength{\arraycolsep}{0.5pt}
\begin{array}{cl}
\mathcal{I}_{e_{a,i}}&=\text{QUREY}(\mathcal{T},e_{a,i}',k)
\label{query_KDtree}
\end{array}
\end{equation}

\noindent \textbf{Selecting optimal trigger} We replace the token $a_{i}$ with the $j$-th token in $\mathcal{I}_{e_{a,i}}$ to generate candidate triggers and employ Equation (\ref{construct_D'}) to create the adversarial dataset $\mathcal{D}_{a'}$. Subsequently, we calculate the loss $l'$ of the dataset $\mathcal{D}_{a'}$ using Equation (\ref{loss_l}). We iterate through each candidate token in $\mathcal{I}_{e_{a,i}}$ (i.e. $j=1,2,...,k$) and employ a greedy strategy to choose the token that minimizes the loss, thereby replacing the original token. 

We iteratively execute the entire attack process until either the optimal triggers $a^*$ are found or the maximum number of search rounds $N$ is reached. The algorithm's complexity is $\mathcal{O}(N \times K \times n_{a} \times k)$.

\begin{algorithm}[htb]
    \setlength{\abovedisplayskip}{3pt plus 2pt minus 18pt}
    \setlength{\belowdisplayskip}{3pt plus 2pt minus 18pt}
    \caption{Attack Process}
    \label{algorithm: Attack Process}
    \begin{algorithmic}[1]
        % 定义输入输出
        \REQUIRE‌ initialized triggers $a$, attack mode $\mathcal{M}=\{\mathcal{M}_{nt},\mathcal{M}_{t}\}$, maximum search rounds $N$.
        \ENSURE‌ the optimal triggers $a^*$
        
        \vspace*{0.5\baselineskip}  %空行

        % 初始化
        % 用(\ref{create_KDtree})创建构建KD-tree来得到$\mathcal{T}$；
        \STATE create a KD-tree with (\ref{create_KDtree}) to get $\mathcal{T}$

        \FOR{$epoch = 0$ to $N$}
            % 获取梯度：构建对抗样本数据集、计算loss和计算梯度
            % 用(\ref{construct_D'})构建数据集来得到$\mathcal{D}_{a}$;
            \STATE construct dataset with (\ref{construct_D'}) to get $\mathcal{D}_{a}$
            % 用(\ref{loss_l})在攻击模式$\mathcal{M}$下来得到$l$;
            \STATE calculate loss with (\ref{loss_l}) under $\mathcal{M}$ to get $l$
            % 用(\ref{calculate_gradient})计算梯度来获得$g$;
            \STATE obtain gradients with (\ref{calculate_gradient}) to get $g$

            \FOR{$i = 0$ to $K \times n_{a}$}
                % 收集候选token：梯度下降,查询
                % 用(\ref{GD})计算$e_{a}'$;
                \STATE obtain $e_{a,i}'$ with (\ref{GD})
                % 用(\ref{query_KDtree})查询候选token来获得$\mathcal{I}_{e_{a,i}}$
                \STATE query candidate tokens with (\ref{query_KDtree}) to get $\mathcal{I}_{e_{a,i}}$

                \FOR{$j = 0$ to $k$}
                    % 选择最优候选token：更新trigger、构建数据集、计算loss,对比loss
                    % 用(\ref{candidate_triggers_a'})替换第i个token $a_{i}'$来获得$a'$;
                    % replace the i-th token $a_{i}'$ with (\ref{candidate_triggers_a'}) to get $a'$ \;
                    \STATE $a'=a$       %这里表示重置trigger,
                    \STATE $a_{i}'=\mathcal{I}_{e_{a,i},j}$  %这里表示替换trigger
                    % 用(\ref{construct_D'})构建数据集来得到$\mathcal{D}_{a'}$;
                    \STATE construct dataset with (\ref{construct_D'}) to get $\mathcal{D}_{a'}$
                    % 用(\ref{loss_l})在攻击模式$\mathcal{M}$下来得到$l$;
                    \STATE calculate loss with (\ref{loss_l}) under $\mathcal{M}$ to get $l'$
                    
                    % 对比loss,并更新l和a。如果$l'$小于$l$,则$a=a'$
                    \IF{$l' < l$}
                        \STATE $l=l'$ 
                        \STATE $a=a'$ 
                    \ENDIF
                \ENDFOR
            \ENDFOR
        \ENDFOR
    \end{algorithmic}
\end{algorithm}

\section{Details of Experimental Setup}
\subsection{Dataset Detials}
\label{appendix: Dataset Detials}
% Movie、FEVER、MultiRC的引用删除,已经包含在ERASER中,用来腾出空间。
\noindent \textbf{Movie} involves a binary sentiment classification task, where the goal is to predict the sentiment category based on a given movie review document. 

\noindent \textbf{FEVER} is a binary fact verification dataset where the input consists of a claim and a document, and the task is to determine whether the document supports or refutes the claim. 

\noindent \textbf{MultiRC} is a multiple-choice question answering dataset, which includes a question, a document, and multiple answer options. Each candidate answer is concatenated with the question and validated for correctness using the document. 

\noindent \textbf{Beer and Hotel} Beer \citep{6413815} and Hotel \citep{10.1145/1835804.1835903} are multi-aspect sentiment classification datasets. For our experiments, we specifically focus on the \emph{appearance} aspect for Beer and the \emph{location} aspect for Hotel. 

Table \ref{table: Overview of the dataset} provides details on the type of human-annotated rationale, number of samples, average length, and sparsity.

\begin{table}[h]       
    \centering
    \renewcommand{\arraystretch}{1.2}
    \caption{Overview of five datasets. \emph{Level} is whether the human-annotated rationale is sentence-level (S) or token-level (T), \emph{Tokens} is the average number of tokens in each document, and \emph{Sparsity} is the test dataset's human-annotated rationale percentage (\%).} 
    \label{table: Overview of the dataset}
    \begin{center}
        \resizebox{0.6\columnwidth}{!}{
            \begin{tabular}{p{1.5cm}<{\centering}|p{1cm}<{\centering}|p{3cm}<{\centering}|p{1.5cm}<{\centering}|p{1.5cm}<{\centering}}
                \hline
                Dataset & Level & Size (train/dev/test) & Tokens & Sparsity \\
                \hline
                Movie  & T & 1599\ /\ 200\ /\ 200 & 776.6 & 19.7 \\
                \hline
                FEVER & S & 97755\ /\ 6096\ /\ 6075 & 285.7 & 22.1 \\
                \hline
                MultiRC & S & 24010\ /\ 3214\ /\ 4841 & 300.1 & 19.4 \\
                \hline
                Beer & T & 33782\ /\ 8731\ /\ 936 & 143.2 & 18.3 \\
                \hline
                Hotel & T & 14457\ /\ 1813\ /\ 199 & 154.8 & 8.5 \\
                \hline
            \end{tabular}
        }
    \end{center}
\end{table}

\subsection{Model Detials}
\label{appendix: Model Detials}

% \noindent \textbf{Training objectives and sparsity regularization} The training objectives are presented below:
% \begin{equation}  
%     \label{RNP's Objective}
%     \begin{array}{cl}
%         \mathop{\min}\limits_{\theta_{R},\theta_{C}} &  - \log p(y|x) + \lambda \Omega(m,\mathcal{S})  \ , \\
%         where & \Omega(m,\mathcal{S}) = |\frac{||m||}{L}-\mathcal{S}|
%         % \vspace{-2pt}
%     \end{array}
% \end{equation}
% where $||\cdot||$ represents the $l_{1}$ nrom, $| \cdot |$ denotes the absolute value, and the second term encourages the selection of tokens as rationale to be close to the predefined level $\mathcal{S}$.

\noindent \textbf{RNP} The unsupervised training objective with the sparsity regularization are presented below:
\begin{equation}  
    \label{RNP's Objective - unsupervision}
    \begin{array}{cl}
        l_{RNP} = - \log p(y|x) + \lambda |\frac{||m||}{L}-\mathcal{S}| 
        % \vspace{-2pt}
    \end{array}
\end{equation}

where $||\cdot||$ represents the $l_{1}$ norm, $| \cdot |$ denotes the absolute value, and the second term encourages the selection of tokens as rationale to be close to the predefined level $\mathcal{S}$. 
The supervised training objective with human-annotation rationale $\hat{m}$ is shown below:
\begin{equation} 
    \label{RNP's Objective - Supervision}
    \begin{array}{cl}
        l_{RNP\text{-}sup} = - \log p(y|x) + \lambda |\frac{||m||}{L}-\mathcal{S}| - \gamma \sum \limits_{i} \hat{m}_{i} \log p(m_{i}|x) 
        % \vspace{-2pt}
    \end{array}
\end{equation}

\noindent \textbf{VIB} During training, VIB parameterizes a relaxed Bernoulli distribution $p(m|x)=RelaxedBernoulli(s)$ using token-level (or sentence-level) logits $s \in \mathbb{R}^{L}$. A soft mask $m_{soft}$ is sampled from this distribution, where each token's mask $m_{soft,i}=\sigma(\frac{\log s_{i}+g}{\tau})$ is independently sampled. Here, $\sigma$ represents the sigmoid function, and $g$ denotes the sampled Gumbel noise. The following objectives are optimized:
\begin{equation}  
    \label{VIB's Objective}
    \begin{array}{cl}
        l_{VIB} = - \log p(y|x) + \lambda \text{KL}[p(m|x)||r(m)]
        % \vspace{-2pt}
    \end{array}
\end{equation}
where the second term utilizes a discrete information bottleneck constraint to control the sparsity of rationales, with $r(m)$ representing a predefined prior distribution determined by sparsity $\mathcal{S}$. During the inference stage, VIB directly applies the sigmoid function to produce logits $s$, selecting the top-$k$ tokens (or sentences) with the largest logits as $m_{i}=\mathbbm{1}[s_{i} \in top\text{-}k(s)]$. Similarly, $k$ is determined by the sparsity $\mathcal{S}$.

The objectives of supervised training with human-annotation rationale $\hat{m}$ are as follows:
\begin{equation}  
    \label{VIB's Objective - Supervision}
    \begin{array}{cl}
        l_{VIB\text{-}sup} = - \log p(y|x) + \lambda \text{KL}[p(m|x)||r(m)] - \gamma \sum \limits_{i} \hat{m}_{i} \log p(m_{i}|x) 
        % \vspace{-2pt}
    \end{array}
\end{equation}

\noindent \textbf{SPECTRA} The predefined constraints of the factor graph are implemented using the $score(\cdot)$ function, enabling the model to generate differentiable binary masks. SPECTRA employs the $BUDGET$ factor, which requires the number of ``1''s in the binary mask to be less than the predefined parameter $B$, i.e., $\sum_{k}z_{k}\le B$, where $B=L \times \mathcal{S}$. Because the entire computation is deterministic, it can be backpropagated using the LP-SparseMAP solver.
\begin{equation}  
    \label{SPECTRA's Objective}
    \begin{array}{cl}
        l_{SPECTRA} = - \log p(y|x) \\
        m = \arg\max_{m'\in \{0,1\}^{L}} (\text{score}(m';\mathcal{S}) - \frac{1}{2}||m'||^2)    
        % \vspace{-2pt}
    \end{array}
\end{equation}
The second term of $m$ encourages the binary mask to be as sparse as possible.
For supervised training with human-annotation rationale $\hat{m}$, the optimization formulation is as follows:
\begin{equation}  
    \label{SPECTRA's Objective - Supervision}
    \begin{array}{cl}
        l_{SPECTRA\text{-}sup} = - \log p(y|x) - \gamma \sum \limits_{i} \hat{m}_{i} \log p(m_{i}|x) \\
        m = \arg\max_{m'\in \{0,1\}^{L}} (\text{score}(m';\mathcal{S}) - \frac{1}{2}||m'||^2)    
        % \vspace{-2pt}
    \end{array}
\end{equation}

\noindent \textbf{FR and DR} FR employs a single encoder for both the rationalizer and predictor, sampling token-level (or sentence-level) logits $s$ using Gumbel-Softmax. DR adjusts the rationalizer's learning rate dynamically in each epoch based on Lipschitz continuity to ensure stable model training. The sparsity constraints in FR and DR are consistent with RNP. Both FR and DR are trained unsupervisedly using Equation (\ref{RNP's Objective - unsupervision}), and supervisedly using Equation (\ref{RNP's Objective - Supervision}) for human-annotated rationales.

\subsection{More Training Details}
\label{appendix: More Training Details} 
In our experiments, each model is implemented using both BERT and Bi-GRU as encoders. For both encoders, we apply an early stopping strategy during training. The maximum input length is 256 tokens. For BERT, we vary the learning rates from 5e-6 to 5e-5 and the batch sizes from 8 to 32. For GRU, we explore learning rates from 1e-5 to 1e-3 and batch sizes from 128 to 512. The optimal learning rate and batch size are determined based on the task's performance on the development set. Details of the model parameters are provided in Tables \ref{Table: BERT's Hyperparameters}.

\noindent \textbf{Input formats} When BERT is employed as the encoder, we initialize the rationalizer and predictor with pre-trained BERT \citep{devlin-etal-2019-bert}, while freezing the first 10 layers. The input format is ``[[CLS], Query, [SEP], Context, [SEP]]''. If a dataset does not include the ``query'' field, we omit both the query and the first ``[SEP]''. When Bi-GRU is used as the encoder, we utilize 100-dimensional Glove embeddings and a Bi-GRU with a hidden state dimension of 200. The input format is ``[Query,Context]''. Following the approach of \citet{chen-etal-2022-rationalization}, for token-level rationale, the format is :
\begin{equation*}
    \setlength{\abovedisplayskip}{3pt plus 2pt minus 8pt}
    \setlength{\belowdisplayskip}{3pt plus 2pt minus 8pt}
    \text{BERT}(x)= (h_{[CLS]},h_{0}^{1},h_{0}^{2},...,h_{0}^{n_0}, \\
     h_{[SEP]},h_{1}^{1},h_{1}^{2},...,h_{1}^{n_1},..., \\
     h_{T}^{1},h_{T}^{2},...,h_{T}^{n_T},h_{[SEP]})
\end{equation*}
where the query is indexed as 0, and the context is indexed from 1 to T. For sentence-level rationale, we concatenate the start and end vectors of each sentence, resulting in a representation in the form of $h_{t} = [ h_{t,0} ; h_{t,n_{t}}]$ for the $t$-th sentence. These representations are then passed through a linear layer to generate logits $w^T h_{t} + b$.

%BERT和GRU作为编码器的参数
\begin{table}[h]       
    \renewcommand{\arraystretch}{1.2}   %设置每一行内容上下方的空间
    \setlength{\tabcolsep}{3pt}    %设置列之间的距离
    \centering
    % 采用BERT作为编码器的超参数的表格。
    \caption{Hyperparameters for Using BERT or GRU as the Encoder}
    \label{Table: BERT's Hyperparameters}
    \begin{center}
        \begin{small}
            \resizebox{0.65\linewidth}{!}{
                \begin{tabular}{p{4cm}<{\centering}|ccccc|ccccc}
                    \hline
                    % 表头和指标
                    & \multicolumn{5}{c|}{BERT} & \multicolumn{5}{c}{GRU} \\
                    \cline{2-11} 
                    & RNP & VIB & SPECTRA & FR & DR & RNP & VIB & SPECTRA & FR & DR  \\
                    \hline 
                    % Learning Rate & 1e-5 & 1e-5 & 1e-5 & 1e-5 & 1e-5 & 1e-4 & 1e-4 & 1e-4 & 1e-4 & 1e-4 \\
                    % \hline 
                    % Batch Size & 16 & 16 & 16 & 16 & 16 & 256 & 256 & 256 & 256 & 256 \\
                    % \hline 
                    Epochs & 450 & 450 & 450 & 450 & 450 & 800 & 800 & 800 & 800 & 800 \\
                    \hline 
                    Hidden Dimension & 768 & 768 & 768 & 768 & 768 & 200 & 200 & 200 & 200 & 200 \\
                    \hline 
                    Frozen Layers & \multicolumn{5}{c|}{Embeddings, Layer 0 \text{-} 9} & - & - & - & - & - \\
                    \hline 
                    Gradient Accumulation Steps & 4 & 4 & 4 & 4 & 4 & - & - & - & - & - \\
                    \hline 
                    Max Gradient Norm & 0.5 & 0.5 & 0.5 & 0.5 & 0.5 & - & - & - & - & - \\
                    \hline 
                    Patient & 20 & 20 & 20 & 20 & 20 & 50 & 50 & 50 & 50 & 50 \\
                    \hline
                    Temperature & 1 & 0.1 & 0.01 & 1 & 1 & 1 & 0.1 & 0.01 & 1 & 1 \\
                    \hline
                    $\lambda$ & 10 & 0.01 & - & 10 & 10 & 10 & 0.01 & - & 10 & 10 \\
                    \hline
                    $\gamma$ & 10 & 1 & 1 & 10 & 10 & 10 & 1 & 1 & 10 & 10 \\
                    \hline
                    Solver Iterate & - & - & 100 & - & - & - & - & 100 & - & - \\
                    \hline
                \end{tabular}	
            }
        \end{small}
    \end{center}
\end{table}

\section{More Experimental Results}       
\label{appendix: More Experimental Results}
% 附录-对比陈丹奇人工手写的攻击文本。 
\noindent \textbf{Comparison with hand-written triggers} Table \ref{Table: Comparison Hand-written Triggers} presents a comparison between our approach and the hand-written triggers proposed by \citet{chen-etal-2022-rationalization}. The hand-written triggers noticeably decrease model performance on the task but have only a minor influence on the quality of the rationales. Particularly, hand-written triggers do not yield a high AR, consistent with the findings of \citet{chen-etal-2022-rationalization}. In contrast, our approach primarily focuses on attacking rationale extraction, resulting in a significant decrease in $GR$ and a higher AR.

\begin{table}[h]       
    \renewcommand{\arraystretch}{1.2}   %设置每一行内容上下方的空间
    \setlength{\tabcolsep}{3pt}    %设置列之间的距离
    \centering
    % 与chen等人构造的trigger对比。我们在beer数据集上对SPECTRA实施攻击。
    \caption{Comparison with the hand-written triggers. We conducted attacks against SPECTRA using the beer dataset. ``Ori\text{-}'' and ``Att\text{-}'' represent the scores before and after the attack, respectively.}
    \label{Table: Comparison Hand-written Triggers}
    \begin{center}
        \begin{small}
            \resizebox{0.8\linewidth}{!}{
                \begin{tabular}{p{4cm}<{\centering}|cccccccccc}
                    \hline
                    % 表头和指标
                    & Ori\text{-}Acc & Att\text{-}Acc & $|\triangle Acc|$ & Ori\text{-}GR & Att\text{-}GR & $\triangle GR$ & $AR$ & $F1_{shift}$ & $F1_{shift,\hat{m}}$ \\
                    \hline 
                    UAT2E & \multirow{2}{*}{\makecell{91.67 $\to$}} & 91.13 & 0.54 & \multirow{2}{*}{\makecell{38.56 $\to$}} & 26.94 & 11.62 & 51.13 & 58.48 & 83.08 \\
                    Hand-Written Triggers & & 76.92 & 14.75 & & 37.04 & 3.95 & 14.91 & 83.29 & 90.74 \\
                    \hline
                \end{tabular}
            }
        \end{small}
    \end{center}
\end{table}

\clearpage
\section{More visualized results}
\label{appendix: More visualized results}

% 展示一些例子：我们提供了攻击前、后的可视化rationale,结果如Table 1-3所示。
We also provide the visualized rationales before and after the attack, as shown in Tables \ref{pick_hotel_GRU}-\ref{pick_multirc_GRU}.

% 【挑选样例 - 横版】
% % hotel - gru - non-target - 117√
\begin{table*}[ht]            %数据集为横轴,方法为纵轴         
    \renewcommand{\arraystretch}{1.5}   %设置每一行内容上下方的空间
    \setlength{\tabcolsep}{3pt}    %设置列之间的距离
    \centering
    \caption{Examples of generated rationales before and after the attack. The RNP, VIB, SPECTRA, FR, and DR models are constructed using \emph{GRU} as the encoder, and attacks are conducted on the \emph{Hotel} dataset. The attack mode is \emph{non-target} attack. Human annotated rationales are \uline{underlined}. Rationales before attacks, triggers and rationales after attacks are highlighted in \textcolor{blue}{blue}, \textcolor{red}{red} and \colorbox{yellow}{yellow}, respectively.}
    \label{pick_hotel_GRU}
    \begin{center}
        \resizebox{0.99\linewidth}{!}{
            \begin{tabular}{p{1.5cm}<{\centering}|L{17.5cm}}
                \hline
                Model & \multicolumn{1}{c}{Context} \\
                \hline
                
                RNP & \textcolor{red}{nauseatingly think even thought} \colorbox{yellow}{\textcolor{red}{first} \uline{\textcolor{blue}{great}}} \uline{location} and great staff made this \colorbox{yellow}{\textcolor{blue}{hotel}} the perfect spot for our roman holiday. the \colorbox{yellow}{\textcolor{blue}{rooms}} on the 6th \colorbox{yellow}{\textcolor{blue}{floor}} have balconies looking out on the pantheon and piazza della \colorbox{yellow}{\textcolor{blue}{rotunda}}. \textcolor{red}{, time this same the the the the the all time this came this the the the as the this} \\
                \hline
                VIB & \textcolor{red}{farming ponds the the} \textcolor{red}{\colorbox{yellow}{local}} \colorbox{yellow}{\textcolor{blue}{\uline{great location}}} and \colorbox{yellow}{\textcolor{blue}{great}} staff \textcolor{blue}{made this} hotel the \colorbox{yellow}{\textcolor{blue}{perfect spot}} for our roman holiday. the rooms on the 6th floor have balconies looking out on the pantheon and piazza della rotunda. \textcolor{red}{place another} \textcolor{red}{\colorbox{yellow}{place}} \textcolor{red}{well the} \textcolor{red}{\colorbox{yellow}{place}} \textcolor{red}{a another. one place another} \textcolor{red}{\colorbox{yellow}{where wetlands here place}} \textcolor{red}{only this the only} \\
                \hline
                SPECTRA & \textcolor{red}{subsequently either well on} \textcolor{red}{\colorbox{yellow}{one}} \colorbox{yellow}{\uline{\textcolor{blue}{great} location}} and \colorbox{yellow}{\textcolor{blue}{great}} staff made this hotel the \colorbox{yellow}{\textcolor{blue}{perfect}} spot for \colorbox{yellow}{our} roman holiday. the rooms on the 6th floor have balconies looking out on the pantheon and piazza della rotunda. \textcolor{red}{least well another and the one well this the same because out the as. part part come part later} \\
                \hline
                FR & \textcolor{red}{both. made however both} \uline{\textcolor{blue}{great} location} and \textcolor{blue}{great} staff made this hotel the \textcolor{blue}{perfect} spot for our roman holiday. the rooms on the 6th floor have balconies looking out on the pantheon \colorbox{yellow}{and} piazza della \textcolor{blue}{rotunda}. \textcolor{red}{both the well made both time all as full instead all though the all all came well made full the} \\  
                \hline
                DR & \textcolor{red}{florida 3,094 as day} \colorbox{yellow}{\textcolor{red}{the}} \colorbox{yellow}{\uline{\textcolor{blue}{great} location}} and great staff \textcolor{blue}{made} this hotel the perfect spot for our roman holiday. the rooms on the 6th floor have \colorbox{yellow}{balconies} looking \textcolor{blue}{out} on the pantheon \colorbox{yellow}{and} piazza della \textcolor{blue}{rotunda}. \textcolor{red}{time followed followed well the though time tuesday the k977-1 preceded made made most the as same made the the}\\
                \hline  
                \multicolumn{2}{r}{Label: Positive, Prediction: Positive} \\  
                \hline
            \end{tabular}	
        }
    \end{center}
\end{table*}
% fever - BERT - Target - 161091 √
\begin{table*}[ht]            %数据集为横轴,方法为纵轴     
    \renewcommand{\arraystretch}{1.5}   %设置每一行内容上下方的空间
    \setlength{\tabcolsep}{3pt}    %设置列之间的距离
    \centering
    \caption{Examples of generated rationales before and after the attack. The RNP, VIB, SPECTRA, FR, and DR models are constructed using \emph{BERT} as the encoder, and attacks are conducted on the \emph{FEVER} dataset. The attack mode is \emph{target} attack. Human annotated rationales are \uline{underlined}. Rationales before attacks, triggers and rationales after attacks are highlighted in \textcolor{blue}{blue}, \textcolor{red}{red} and \colorbox{yellow}{yellow}, respectively.}
    \label{pick_fever_BERT}
    \begin{center}
        \resizebox{0.99\linewidth}{!}{
            \begin{tabular}{p{1.5cm}<{\centering}|L{17.5cm}}
                \hline
                Model & \multicolumn{1}{c}{Context} \\
                \hline
                    
                RNP & robert singer ( producer ) has spent his entire career producing only live theater.  \textcolor{red}{uno kimZh ! chev} robert singer is an american television producer, director and writer. \textcolor{blue}{\uline{he is best known for his work on supernatural where he serves as an executive producer, director and occasional writer.}} \textcolor{red}{! ags ! ! !} \colorbox{yellow}{the character bobby singer was named after him.} a fictionalized version of singer appears in the season 6 episode the french mistake played by brian doyle - murray. \colorbox{yellow}{\textcolor{red}{mu ! en mikhail kong}} \uline{singer also created the series reasonable doubts and has worked as a producer and director on various television series, including lois $\&$ clark : the new adventures of superman.} \textcolor{blue}{\uline{he also served as a producer for the film cujo}} \textcolor{red}{vna ! of ! \text{ye} laurence ! ! ! !} \\
                \hline
                VIB &  robert singer ( producer ) has spent his entire career producing only live theater.  \colorbox{yellow}{\textcolor{red}{situation bedfordLoe 1/2 '}} \colorbox{yellow}{\textcolor{blue}{robert singer is an american television producer, director and writer.}} \textcolor{blue}{\uline{he is best known for his work on supernatural where he serves as an executive producer, director and occasional writer.}} \colorbox{yellow}{\textcolor{red}{nec ' 1016 mw '}} the character bobby singer was named after him. a fictionalized version of singer appears in the season 6 episode the french mistake played by brian doyle - murray. \textcolor{red}{! @ 1016 bedford both} \uline{singer also created the series reasonable doubts and has worked as a producer and director on various television series, including lois $\&$ clark : the new adventures of superman. he also served as a producer for the film cujo} \textcolor{red}{mani qaeda icao l xiv immunity ` ! ! !} \\
                \hline
                SPECTRA &  robert singer ( producer ) has spent his entire career producing only live theater.  \colorbox{yellow}{\textcolor{red}{en ! ored ! precedent}} \colorbox{yellow}{\textcolor{blue}{robert singer is an american television producer, director and writer.}} \uline{he is best known for his work on supernatural where he serves as an executive producer, director and occasional writer.} \textcolor{red}{oom ! vity ! culture} the character bobby singer was named after him. a fictionalized version of singer appears in the season 6 episode the french mistake played by brian doyle - murray. \textcolor{red}{lain ! ! atRill} \textcolor{blue}{\uline{singer also created the series reasonable doubts and has worked as a producer and director on various television series, including lois $\&$ clark : the new adventures of superman.} he also served as a producer for the film cujo} \colorbox{yellow}{\textcolor{red}{though ! ! ! !}} \textcolor{red}{! ! kv ! some} \\
                \hline
                FR &  robert singer ( producer ) has spent his entire career producing only live theater.  \colorbox{yellow}{\textcolor{red}{bled prize refereesTility optics}} robert singer is an american television producer, director and writer. \colorbox{yellow}{\uline{he is best known for his work on supernatural where he serves as an executive producer, director and occasional writer.}} \colorbox{yellow}{\textcolor{red}{rraCation cfl qin ‡}} \colorbox{yellow}{the character bobby singer was named after him.} \textcolor{blue}{a fictionalized version of singer appears in the season 6 episode the french mistake played by brian doyle - murray.} \colorbox{yellow}{\textcolor{red}{rce subspRator fey reversal}} \uline{singer also created the series reasonable doubts and has worked as a producer and director on various television series, including lois $\&$ clark : the new adventures of superman.} \textcolor{blue}{\uline{he also served as a producer for the film cujo}} \colorbox{yellow}{\textcolor{red}{rter recorder lenses. reversalRricular obstruction emmy feminineTility}} \\
                \hline
                DR & robert singer ( producer ) has spent his entire career producing only live theater.  \colorbox{yellow}{\textcolor{red}{yraAast ne ! vn}} \textcolor{blue}{robert singer is an american television producer, director and writer. \uline{he is best known for his work on supernatural where he serves as an executive producer, director and occasional writer.}} \colorbox{yellow}{\textcolor{red}{rraCation cfl qin ‡} the character bobby singer was named after him.} a fictionalized version of singer appears in the season 6 episode the french mistake played by brian doyle - murray. \textcolor{red}{! ! ! / !} \textcolor{blue}{\uline{singer also created the series reasonable doubts and has worked as a producer and director on various television series, including lois $\&$ clark : the new adventures of superman.}} \uline{he also served as a producer for the film cujo} \colorbox{yellow}{\textcolor{red}{nelly ! ! ! veredMmAashi ! ! !}}\\  
                \hline 
                \multicolumn{2}{r}{Label: Negative, Prediction: Negative} \\  
                \hline
            \end{tabular}	
        }
    \end{center}
\end{table*}
% fever - GRU - Target - 86783 √
\begin{table*}[ht]            %数据集为横轴,方法为纵轴           
    \renewcommand{\arraystretch}{1.5}   %设置每一行内容上下方的空间
    \setlength{\tabcolsep}{3pt}    %设置列之间的距离
    \centering
    \caption{Examples of generated rationales before and after the attack. The RNP, VIB, SPECTRA, FR, and DR models are constructed using \emph{GRU} as the encoder, and attacks are conducted on the \emph{FEVER} dataset. The attack mode is \emph{target} attack. Human annotated rationales are \uline{underlined}. Rationales before attacks, triggers and rationales after attacks are highlighted in \textcolor{blue}{blue}, \textcolor{red}{red} and \colorbox{yellow}{yellow}, respectively.}
    \label{pick_fever_GRU}
    \begin{center}
        \resizebox{0.99\linewidth}{!}{
            \begin{tabular}{p{1.5cm}<{\centering}|L{17.5cm}}
                \hline
                Model & \multicolumn{1}{c}{Context} \\
                \hline
                    
                RNP & Ned Stark's cast was nominated for an Academy Award. \colorbox{yellow}{\textcolor{red}{enormousness usanee e. explains}} Eddard `` Ned '' Stark is a fictional character in the first book of George R. R. Martin's A Song of Ice and Fire series of fantasy novels and the first season of its television adaptation. Introduced in 1996's A Game of Thrones, Ned is the honorable lord of Winterfell, an ancient fortress in the North of the fictional continent of Westeros. \textcolor{red}{boosting soullessness sure fact.} Though the character is established as a primary character in the novel and the TV adaptation, Martin's plot twist at the end involving Ned shocked both readers of the book and viewers of the TV series. \textcolor{blue}{Ned is portrayed by Sean Bean in the first season of HBO's adaptation of the series, Game of Thrones, and as a child and young man by Sebastian Croft and Robert Aramayo respectively in the sixth season.} \colorbox{yellow}{\textcolor{red}{the acknowledging bigotries convince presumably}} Bean was nominated for a Saturn Award for Best Actor on Television and a Scream Award for Best Fantasy Actor for the role. \textcolor{blue}{\uline{He and the rest of the cast were nominated for Screen Actors Guild Awards for Outstanding Performance by an Ensemble in a Drama Series in 2011.}} \colorbox{yellow}{\textcolor{red}{concerns ask, the moreover seventh. 1977-2010 rongfeng once}}  \\
                \hline
                VIB & Ned Stark's cast was nominated for an Academy Award. \textcolor{red}{and being both of time} Eddard `` Ned '' Stark is a fictional character in the first book of George R. R. Martin's A Song of Ice and Fire series of fantasy novels and the first season of its television adaptation. Introduced in 1996's A Game of Thrones, Ned is the honorable lord of Winterfell, an ancient fortress in the North of the fictional continent of Westeros. \textcolor{red}{. from too that time} Though the character is established as a primary character in the novel and the TV adaptation, Martin's plot twist at the end involving Ned shocked both readers of the book and viewers of the TV series. \textcolor{blue}{Ned is portrayed by Sean Bean in the first season of HBO's adaptation of the series, Game of Thrones, and as a child and young man by Sebastian Croft and Robert Aramayo respectively in the sixth season.} \colorbox{yellow}{\textcolor{red}{and some of in time}} Bean was nominated for a Saturn Award for Best Actor on Television and a Scream Award for Best Fantasy Actor for the role. \textcolor{blue}{\uline{He and the rest of the cast were nominated for Screen Actors Guild Awards for Outstanding Performance by an Ensemble in a Drama Series in 2011.}} \colorbox{yellow}{\textcolor{red}{and but and in where putting. same time so}} \\
                \hline 
                SPECTRA & Ned Stark's cast was nominated for an Academy Award. \textcolor{red}{. over another over as} Eddard `` Ned '' Stark is a fictional character in the first book of George R. R. Martin's A Song of Ice and Fire series of fantasy novels and the first season of its television adaptation. Introduced in 1996's A Game of Thrones, Ned is the honorable lord of Winterfell, an ancient fortress in the North of the fictional continent of Westeros. \textcolor{red}{. 1970s on made expected} Though the character is established as a primary character in the novel and the TV adaptation, Martin's plot twist at the end involving Ned shocked both readers of the book and viewers of the TV series. Ned is portrayed by Sean Bean in the first season of HBO's adaptation of the series, Game of Thrones, and as a child and young man by Sebastian Croft and Robert Aramayo respectively in the sixth season. \colorbox{yellow}{\textcolor{red}{. put off on 54}} Bean was nominated for a Saturn Award for Best Actor on Television and a Scream Award for Best Fantasy Actor for the role. \textcolor{blue}{\uline{He and the rest of the cast were nominated for Screen Actors Guild Awards for Outstanding Performance by an Ensemble in a Drama Series in 2011.}} \colorbox{yellow}{\textcolor{red}{.. made made an.... only}}  \\
                \hline
                FR & \colorbox{yellow}{Ned Stark's cast was nominated for an Academy Award. \textcolor{red}{'' ny455 also of a}} Eddard `` Ned '' Stark is a fictional character in the first book of George R. R. Martin's A Song of Ice and Fire series of fantasy novels and the first season of its television adaptation. Introduced in 1996's A Game of Thrones, Ned is the honorable lord of Winterfell, an ancient fortress in the North of the fictional continent of Westeros. \textcolor{red}{whatever indeed the one west-southwestward} Though the character is established as a primary character in the novel and the TV adaptation, Martin's plot twist at the end involving Ned shocked both readers of the book and viewers of the TV series. Ned is portrayed by Sean Bean in the first season of HBO's adaptation of the series, Game of Thrones, and as a child and young man by Sebastian Croft and Robert Aramayo respectively in the sixth season. \colorbox{yellow}{\textcolor{red}{rationale including the the unreviewed}} Bean was nominated for a Saturn Award for Best Actor on Television and a Scream Award for Best Fantasy Actor for the role. \textcolor{blue}{\uline{He and the rest of the cast were nominated for Screen Actors Guild Awards for Outstanding Performance by an Ensemble in a Drama Series in 2011.}} \colorbox{yellow}{\textcolor{red}{whomever having the the the 1997 the the the well}} \\
                \hline
                DR & Ned Stark's cast was nominated for an Academy Award. \textcolor{red}{\colorbox{yellow}{renewed well :. mukuro}} \textcolor{blue}{\colorbox{yellow}{Eddard `` Ned '' Stark is a fictional character in the first book of George} \colorbox{yellow}{R.R. Martin's A Song of Ice and Fire series of fantasy novels and the first season of its television adaptation.}} Introduced in 1996's A Game of Thrones, Ned is the honorable lord of Winterfell, an ancient fortress in the North of the fictional continent of Westeros. \textcolor{red}{the the the the the} Though the character is established as a primary character in the novel and the TV adaptation, Martin 's plot twist at the end involving Ned shocked both readers of the book and viewers of the TV series. Ned is portrayed by Sean Bean in the first season of HBO 's adaptation of the series, Game of Thrones, and as a child and young man by Sebastian Croft and Robert Aramayo respectively in the sixth season. \textcolor{red}{the the the on the} Bean was nominated for a Saturn Award for Best Actor on Television and a Scream Award for Best Fantasy Actor for the role. \textcolor{blue}{\uline{He and the rest of the cast were nominated for Screen Actors Guild Awards for Outstanding Performance by an Ensemble in a Drama Series in 2011.}} \colorbox{yellow}{\textcolor{red}{the... confronts essayist ;, ex-lover turned}}\\  
                \hline  
                \multicolumn{2}{r}{Label: Positive, Prediction: Positive} \\  
                \hline
            \end{tabular}	
        }
    \end{center}
\end{table*}
% multirc - BERT - target - News_CNN_cnn-3b3301b2f6a5c4974baf18b4cf109206829dbc29.txt:3:0 √
\begin{table*}[ht]            %数据集为横轴,方法为纵轴           
    \renewcommand{\arraystretch}{1.5}   %设置每一行内容上下方的空间
    \setlength{\tabcolsep}{3pt}    %设置列之间的距离
    \centering
    \caption{Examples of generated rationales before and after the attack. The RNP, VIB, SPECTRA, FR, and DR models are constructed using \emph{BERT} as the encoder, and attacks are conducted on the \emph{MultiRC} dataset. The attack mode is \emph{target} attack. Human annotated rationales are \uline{underlined}. Rationales before attacks, triggers and rationales after attacks are highlighted in \textcolor{blue}{blue}, \textcolor{red}{red} and \colorbox{yellow}{yellow}, respectively.}
    \label{pick_multirc_BERT}
    \begin{center}
        \resizebox{0.99\linewidth}{!}{
            \begin{tabular}{p{1.5cm}<{\centering}|L{17.5cm}}
                \hline
                Model & \multicolumn{1}{c}{Context} \\
                \hline
                
                RNP & does the screenroom interview international icons of the film industry such as screenwriter mike leigh ? | | no interview  \colorbox{yellow}{\textcolor{red}{mm cyber discomfort breast korea}} \textcolor{blue}{\uline{london, england ( cnn ) - - british filmmaker, screenwriter and playwright, mike leigh has been in the movie business for over 35 years.}} \colorbox{yellow}{leigh on screenwriting : " i don't make a conventional screenplay... it's a whole organic process. " \textcolor{red}{ented karate scrolling she nippon}} in that time, he has been nominated for five oscars, as well as winning the best director award at cannes film festival for " naked " in 1993, the palme d ' or there for " secrets $\&$ lies " in 1996 and the leono d ' oro at venice film festival in 2004 for " vera drake. " \colorbox{yellow}{famous for his fierce independence ( read refusal to work in hollywood ), leigh ' s work is known for gritty realism and a focus on underprivileged sections} \colorbox{yellow}{of british society.} \colorbox{yellow}{\textcolor{red}{ens ) whom whom cheung}} \textcolor{blue}{another of leigh ' s calling cards is an unconventional approach to screenwriting. "} \uline{the screening room " caught up with the veteran director at the international screenwriters' festival in the uk earlier this year to ask him more about his approach to making films .} \colorbox{yellow}{\textcolor{red}{ede countdownRgan theGuard}} \uline{the screening room : why is this festival so important to you ?} \textcolor{blue}{mike leigh : i am a filmmaker who is both} \textcolor{red}{rteSz rosa erotic govt} \\
                \hline
                VIB &  does the screenroom interview international icons of the film industry such as screenwriter mike leigh ? | | no interview  \colorbox{yellow}{\textcolor{red}{often criticized : traditional philippines}} \uline{london, england ( cnn ) - - british filmmaker, screenwriter and playwright, mike leigh has been in the movie business for over 35 years.} leigh on screenwriting : " i don't make a conventional screenplay... it's a whole organic process. " \textcolor{red}{rooney saga ! s doin} in that time, he has been nominated for five oscars, as well as winning the best director award at cannes film festival for " naked " in 1993, the palme d ' or there for " secrets $\&$ lies " in 1996 and the leono d ' oro at venice film festival in 2004 for " vera drake. " \colorbox{yellow}{\textcolor{blue}{famous for his fierce independence ( read refusal to work in hollywood ), leigh's work is known for gritty realism and a focus on underprivileged sections}} \colorbox{yellow}{\textcolor{blue}{of british society.} \textcolor{red}{alongside professor ( however,}} another of leigh ' s calling cards is an unconventional approach to screenwriting. " \textcolor{blue}{\uline{the screening room " caught up with the veteran director at the international screenwriters' festival in the uk earlier this year to ask him more about his approach to making films .}} \textcolor{red}{! | chico raulEma} \uline{the screening room : why is this festival so important to you ?} mike leigh : i am a filmmaker who is both \textcolor{red}{| cheney couldn hr quotes} \\
                \hline
                SPECTRA &  does the screenroom interview international icons of the film industry such as screenwriter mike leigh ? | | no interview  \textcolor{red}{russianLto presidentEDb} \uline{london, england ( cnn ) - - british filmmaker, screenwriter and playwright, mike leigh has been in the movie business for over 35 years.} leigh on screenwriting : " i don't make a conventional screenplay... it's a whole organic process. " \textcolor{red}{allTSworn trillion !} in that time, he has been nominated for five oscars, as well as winning the best director award at cannes film festival for " naked " in 1993, the palme d ' or there for " secrets $\&$ lies " in 1996 and the leono d ' oro at venice film festival in 2004 for " vera drake. " \colorbox{yellow}{\textcolor{blue}{famous for his fierce independence ( read refusal to work in hollywood ), leigh's work is known for gritty realism and a focus on underprivileged sections}} \colorbox{yellow}{\textcolor{blue}{of british society.}} \textcolor{red}{jit ! ave they !} another of leigh ' s calling cards is an unconventional approach to screenwriting. " \textcolor{blue}{\uline{the screening room " caught up with the veteran director at the international screenwriters' festival in the uk earlier this year to ask him more about his approach to making films .}} \textcolor{red}{sample ! conference section consider} \uline{the screening room : why is this festival so important to you ?} mike leigh : i am a filmmaker who is both \colorbox{yellow}{\textcolor{red}{bingo ! anime ! !}} \\
                \hline
                FR &  does the screenroom interview international icons of the film industry such as screenwriter mike leigh ? | | no interview  \colorbox{yellow}{\textcolor{red}{intense resentment increased ¡ ptolemy}} \uline{london, england ( cnn ) - - british filmmaker, screenwriter and playwright, mike leigh has been in the movie business for over 35 years.} \colorbox{yellow}{leigh on screenwriting : " i don't make a conventional screenplay... it's a whole organic process. " \textcolor{red}{cturing several . most pendleton}} in that time, he has been nominated for five oscars, as well as winning the best director award at cannes film festival for " naked " in 1993, the palme d ' or there for " secrets $\&$ lies " in 1996 and the leono d ' oro at venice film festival in 2004 for " vera drake. " \colorbox{yellow}{famous for his fierce independence ( read refusal to work in hollywood ), leigh's work is known for gritty realism and a focus on underprivileged sections} \colorbox{yellow}{of british society. \textcolor{red}{ved himself harshly most decidedly}} another of leigh's calling cards is an unconventional approach to screenwriting. " \uline{the screening room " caught up with the veteran director at the international screenwriters' festival in the uk earlier this year to ask him more about his approach to making films .} \colorbox{yellow}{\textcolor{red}{gging whatever slightest fists pendleton}} \textcolor{blue}{\uline{the screening room : why is this festival so important to you ?} mike leigh : i am a filmmaker who is both} \colorbox{yellow}{\textcolor{red}{ving ¡ face palmsKala}} \\
                \hline
                DR & does the screenroom interview international icons of the film industry such as screenwriter mike leigh ? | | no interview  \textcolor{red}{starring attempting fellowship ! dation} \uline{london,england ( cnn ) - - british filmmaker, screenwriter and playwright, mike leigh has been in the movie business for over 35 years.} \colorbox{yellow}{leigh on screenwriting : " i do n ' t make a conventional screenplay... it ' s a whole organic process. " \textcolor{red}{\#\#rce parish yun ¡ espionage}} in that time, he has been nominated for five oscars, as well as winning the best director award at cannes film festival for " naked " in 1993, the palme d ' or there for " secrets $\&$ lies " in 1996 and the leono d ' oro at venice film festival in 2004 for " vera drake. " famous for his fierce independence ( read refusal to work in hollywood ), leigh ' s work is known for gritty realism and a focus on underprivileged sections of british society. \colorbox{yellow}{\textcolor{red}{\#\#nched internment attempting ¡boarding}} another of leigh's calling cards is an unconventional approach to screenwriting. " \textcolor{blue}{\uline{the screening room " caught up with the veteran director at the international screenwriters' festival in the uk earlier this year to ask him more about his approach to making films .}} \textcolor{red}{ated kioya ofbiology} \textcolor{blue}{\uline{the screening room : why is this festival so important to you ? mike leigh : i am a filmmaker who is both } }\textcolor{red}{ency homosexuality scientology certifiedMana}\\  
                \hline  
                \multicolumn{2}{r}{Label: Negative, Prediction: Negative} \\  
                \hline
            \end{tabular}	
        }
    \end{center}
\end{table*}
% multirc - GRU - target - History-Anthropology_oanc-HistoryMadrid-2.txt:11:3 √
\begin{table*}[ht]            %数据集为横轴,方法为纵轴  
    \renewcommand{\arraystretch}{1.5}   %设置每一行内容上下方的空间
    \setlength{\tabcolsep}{3pt}    %设置列之间的距离
    \centering
    \caption{Examples of generated rationales before and after the attack. The RNP, VIB, SPECTRA, FR, and DR models are constructed using \emph{GRU} as the encoder, and attacks are conducted on the \emph{MultiRC} dataset. The attack mode is \emph{target} attack. Human annotated rationales are \uline{underlined}. Rationales before attacks, triggers and rationales after attacks are highlighted in \textcolor{blue}{blue}, \textcolor{red}{red} and \colorbox{yellow}{yellow}, respectively.}
    \label{pick_multirc_GRU}
    \begin{center}
        \resizebox{0.99\linewidth}{!}{
            \begin{tabular}{p{1.5cm}<{\centering}|L{17.5cm}}
                \hline
                Model & \multicolumn{1}{c}{Context} \\
                \hline
                
                RNP & Where Charles V was born ? || 1530 \textcolor{red}{why one piyanart time re-think} \colorbox{yellow}{\textcolor{blue}{Spain 's Golden Age : Under Ferdinand and Isabella, Spain underwent a dramatic}} \colorbox{yellow}{\textcolor{blue}{transformation.}} In 1492 the royal pair presided over the final conquest over the Moors and discovery of the New World, including the great wealth that feat brought to Spain. \textcolor{red}{did this the the the} \textcolor{blue}{Spain flourished during a Golden Age, a century of Spanish economic and political supremacy in international affairs, accompanied by marvels of art and literature.} Ferdinand and Isabella were consummate Spaniards, committed to the expansion of the crown. \textcolor{red}{the the the the the} \uline{By contrast, their grandson, who assumed the throne in 1516, was born in Flanders in 1500, and Charles I could barely express himself in Spanish.} \colorbox{yellow}{The first of the Habsburgs, he packed his retinue with Burgundian and Flemish nobles.} \textcolor{red}{the the the the the} \textcolor{blue}{\uline{Soon after his arrival in Spain, the young man inherited the title of Holy Roman Emperor, as Charles V. The responsibilities of the crown kept him busy away from the royal residences of Toledo, Segovia, Valladolid, and Madrid.}} \textcolor{blue}{While the monarch was away on one of his many business trips, his increasingly dissatisfied subjects protested violently.} A revolt of the comuneros, or townsmen, broke out in a number of Spanish cities, including Madrid. \textcolor{red}{the the the the the}  \\
                \hline
                VIB & Where Charles V was born ? || 1530 \textcolor{red}{only neither yet quickly putting} Spain 's Golden Age : Under Ferdinand and Isabella, Spain underwent a dramatic transformation. In 1492 the royal pair presided over the final conquest over the Moors and discovery of the New World, including the great wealth that feat brought to Spain. \colorbox{yellow}{\textcolor{red}{so kept next given just}} Spain flourished during a Golden Age, a century of Spanish economic and political supremacy in international affairs, accompanied by marvels of art and literature. Ferdinand and Isabella were consummate Spaniards, committed to the expansion of the crown. \colorbox{yellow}{\textcolor{red}{addition another the the another}} \textcolor{blue}{\uline{By contrast, their grandson, who assumed the throne in 1516, was born in Flanders in 1500, and Charles I could barely express himself in Spanish.}} The first of the Habsburgs, he packed his retinue with Burgundian and Flemish nobles. \colorbox{yellow}{\textcolor{red}{it several especially yet almost}} \uline{Soon after his arrival in Spain, the young man inherited the title of Holy Roman Emperor, as Charles V. The responsibilities of the crown kept him busy away from the royal residences of Toledo, Segovia, Valladolid, and Madrid.} \textcolor{blue}{While the monarch was away on one of his many business trips, his increasingly dissatisfied subjects protested violently.} A revolt of the comuneros, or townsmen, broke out in a number of Spanish cities, including Madrid. \textcolor{red}{another first first part another} \\
                \hline 
                SPECTRA & Where Charles V was born ? || 1530 \colorbox{yellow}{\textcolor{red}{still none put the of}} Spain 's Golden Age : Under Ferdinand and Isabella, Spain underwent a dramatic transformation. In 1492 the royal pair presided over the final conquest over the Moors and discovery of the New World, including the great wealth that feat brought to Spain. \colorbox{yellow}{\textcolor{red}{these namely time one end}} Spain flourished during a Golden Age, a century of Spanish economic and political supremacy in international affairs, accompanied by marvels of art and literature. \textcolor{blue}{Ferdinand and Isabella were consummate Spaniards, committed to the expansion of the crown.} \colorbox{yellow}{\textcolor{red}{particular then then one time}} \textcolor{blue}{\uline{By contrast, their grandson, who assumed the throne in 1516, was born in Flanders in 1500, and Charles I could barely express himself in Spanish.}} The first of the Habsburgs, he packed his retinue with Burgundian and Flemish nobles. \textcolor{red}{what then the the however} \uline{Soon after his arrival in Spain, the young man inherited the title of Holy Roman Emperor, as Charles V. The responsibilities of the crown kept him busy away from the royal residences of Toledo, Segovia, Valladolid, and Madrid.} While the monarch was away on one of his many business trips, his increasingly dissatisfied subjects protested violently. A revolt of the comuneros, or townsmen, broke out in a number of Spanish cities, including Madrid. \textcolor{red}{of of the the out}  \\
                \hline
                FR & Where Charles V was born ? || 1530 \textcolor{red}{baqerabad it the viers hobel} \textcolor{blue}{Spain 's Golden Age : Under Ferdinand and Isabella, Spain underwent a dramatic transformation.} In 1492 the royal pair presided over the final conquest over the Moors and discovery of the New World, including the great wealth that feat brought to Spain. \textcolor{red}{chāy we the the christened} Spain flourished during a Golden Age, a century of Spanish economic and political supremacy in international affairs, accompanied by marvels of art and literature. Ferdinand and Isabella were consummate Spaniards, committed to the expansion of the crown. \textcolor{red}{the that the the newly} \uline{By contrast, their grandson, who assumed the throne in 1516, was born in Flanders in 1500, and Charles I could barely express himself in Spanish.} \colorbox{yellow}{The first of the Habsburgs, he packed his retinue with Burgundian and Flemish nobles.} \textcolor{red}{most of all the personally} \textcolor{blue}{\uline{Soon after his arrival in Spain, the young man inherited the title of Holy Roman Emperor, as Charles V. The responsibilities of the crown kept him busy away from the royal residences of Toledo, Segovia, Valladolid, and Madrid.}} \textcolor{blue}{While the monarch was away on one of his many business trips, his increasingly dissatisfied subjects protested violently.} A revolt of the comuneros, or townsmen, broke out in a number of Spanish cities, including Madrid. \textcolor{red}{this end place assumed.} \\
                \hline
                DR & Where Charles V was born ? || 1530 \colorbox{yellow}{\textcolor{red}{one the the the another} \textcolor{blue}{Spain 's Golden Age : Under Ferdinand and Isabella, Spain underwent a dramatic transformation.}} In 1492 the royal pair presided over the final conquest over the Moors and discovery of the New World, including the great wealth that feat brought to Spain. \textcolor{red}{the the the the the} Spain flourished during a Golden Age, a century of Spanish economic and political supremacy in international affairs, accompanied by marvels of art and literature. Ferdinand and Isabella were consummate Spaniards, committed to the expansion of the crown. \textcolor{red}{the the the the the} \uline{By contrast, their grandson, who assumed the throne in 1516, was born in Flanders in 1500, and Charles I could barely express himself in Spanish.} \colorbox{yellow}{The first of the Habsburgs, he packed his retinue with Burgundian and Flemish nobles.} \textcolor{red}{the the the the the} \textcolor{blue}{\uline{Soon after his arrival in Spain, the young man inherited the title of Holy Roman Emperor, as Charles V. The responsibilities of the crown kept him busy away from the royal residences of Toledo, Segovia, Valladolid, and Madrid.}} \textcolor{blue}{While the monarch was away on one of his many business trips, his increasingly dissatisfied subjects protested violently.} A revolt of the comuneros, or townsmen, broke out in a number of Spanish cities, including Madrid. \textcolor{red}{the the the keep the} \\  
                \hline  
                \multicolumn{2}{r}{Label: Negative, Prediction: Negative} \\  
                \hline
            \end{tabular}	
        }
    \end{center}
\end{table*}

\end{document}